\pdfoutput=1
\documentclass[sigconf]{acmart}

\AtBeginDocument{%
  }

\setcopyright{acmlicensed}
\copyrightyear{2024}
\acmYear{2024}

\acmDOI{10.1145/3664647.3681083}
\acmISBN{979-8-4007-0686-8/24/10}

\acmConference[MM '24] {Proceedings of the 32nd ACM International Conference on Multimedia}{October 28--November 1, 2024}{Melbourne, VIC, Australia.}
\acmBooktitle{Proceedings of the 32nd ACM International Conference on Multimedia (MM '24), October 28--November 1, 2024, Melbourne, VIC, Australia}
\usepackage{subfig}
\usepackage{amsmath}
\usepackage{tabularx}
\usepackage{multirow}
\usepackage{color}
\usepackage{titlesec}
\usepackage{xcolor}
\usepackage{graphicx}
\usepackage{float}
\usepackage{colortbl}
\usepackage{booktabs}  
\usepackage{adjustbox} % For resizebox  
\usepackage{hyperref}
\usepackage{cleveref}
\begin{document}

\title{DMFourLLIE: Dual-Stage and Multi-Branch Fourier Network for Low-Light Image Enhancement}

\settopmatter{authorsperrow=4}

\author{Tongshun Zhang}
\email{tszhang23@mails.jlu.edu.cn}
\affiliation{%
  \institution{Jilin University}
  \department{College of Computer Science and Technology}
  \city{Changchun}
  \country{China}
}

\author{Pingping Liu}
\authornote{Corresponding author.}
\email{liupp@jlu.edu.cn}
\affiliation{%
  \institution{Jilin University}
  \department{College of Computer Science and Technology}
  \city{Changchun}
  \country{China}
}

\author{Ming Zhao}
\email{mingzhao23@mails.jlu.edu.cn}
\affiliation{%
  \institution{Jilin University}
  \department{College of Software}
  \city{Changchun}
  \country{China}
}

\author{Haotian Lv}
\email{lvht22@mails.jlu.edu.cn}
\affiliation{%
  \institution{Jilin University}
  \department{College of Computer Science and Technology}
  \city{Changchun}
  \country{China}
}

\renewcommand{\shortauthors}{Tongshun Zhang, Pingping Liu, Ming Zhao and Haotian Lv}

\begin{abstract}
 In the Fourier frequency domain, luminance information is primarily encoded in the amplitude component, while spatial structure information is significantly contained within the phase component. Existing low-light image enhancement techniques using Fourier transform have mainly focused on amplifying the amplitude component and simply replicating the phase component, an approach that often leads to color distortions and noise issues. In this paper, we propose a Dual-Stage Multi-Branch Fourier Low-Light Image Enhancement (DMFourLLIE) framework to address these limitations by emphasizing the phase component's role in preserving image structure and detail. The first stage integrates structural information from infrared images to enhance the phase component and employs a luminance-attention mechanism in the luminance-chrominance color space to precisely control amplitude enhancement. The second stage combines multi-scale and Fourier convolutional branches for robust image reconstruction, effectively recovering spatial structures and textures. This dual-branch joint optimization process ensures that complex image information is retained, overcoming the limitations of previous methods that neglected the interplay between amplitude and phase. Extensive experiments across multiple datasets demonstrate that DMFourLLIE outperforms current state-of-the-art methods in low-light image enhancement. Our code is available at \href{https://github.com/bywlzts/DMFourLLIE}{\textcolor{purple}{DMFourLLIE}}.
\end{abstract}




\maketitle

\section{Introduction}

\begin{figure}[htbp]
\vspace{-0.6cm}
\centering
\subfloat[First column is input, second to fourth columns represent the outputs of the respective methods in the first stage.]{
\begin{minipage}[b]{0.24\linewidth}
\centering
\includegraphics[width=1\linewidth]{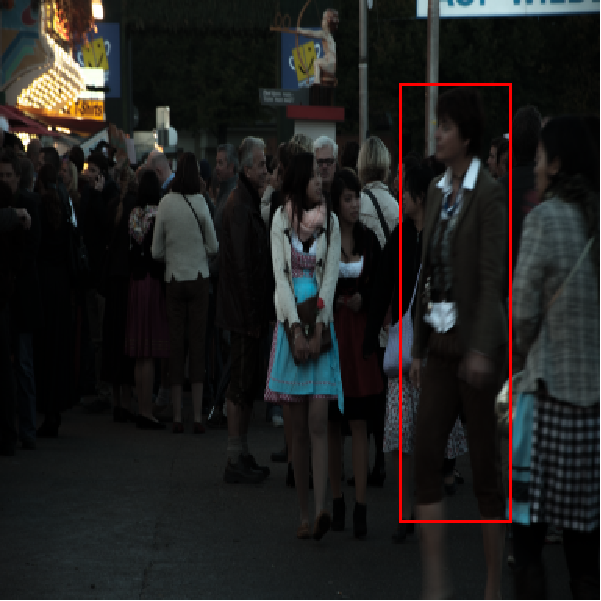}
\vspace{-0.75cm}
\caption*{Input}
\end{minipage}
\begin{minipage}[b]{0.24\linewidth}
\centering
\includegraphics[width=1\linewidth]{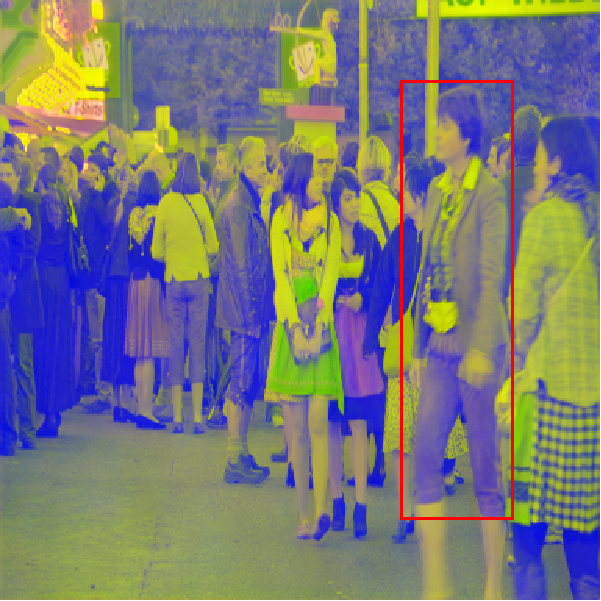}
\vspace{-0.75cm}
\caption*{FECNet}
\end{minipage}
\begin{minipage}[b]{0.24\linewidth}
\centering
\includegraphics[width=1\linewidth]{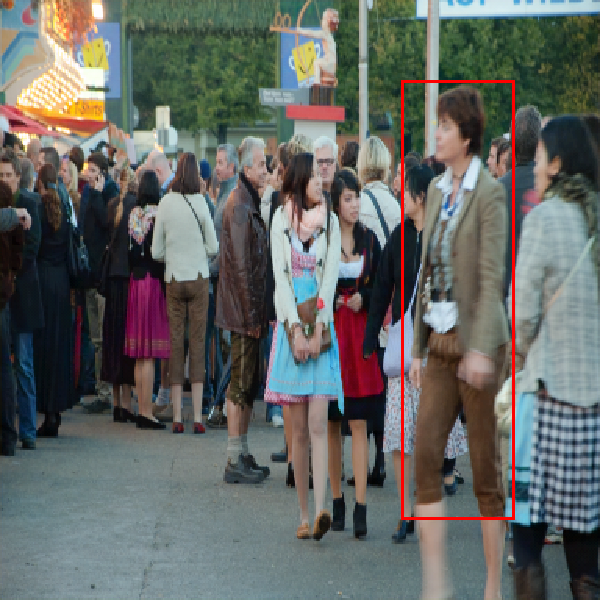}
\vspace{-0.75cm}
\caption*{FourLLIE}
\end{minipage}
\begin{minipage}[b]{0.24\linewidth}
\centering
\includegraphics[width=1\linewidth]{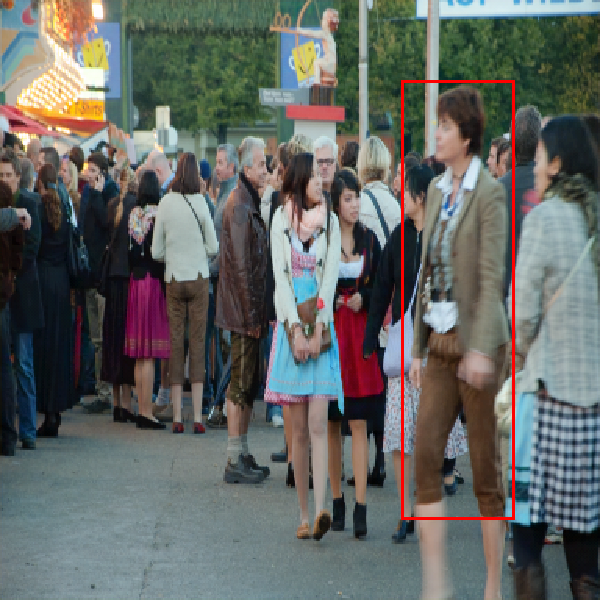}  
\vspace{-0.75cm}
\caption*{DMFourLLIE}
\end{minipage}
\label{fourbasedcomparison-a}
}

\subfloat[First column is ground truth image, second to fourth columns represent the outputs of the respective methods in the second stage.]{
\begin{minipage}[b]{0.24\linewidth}
\centering
\includegraphics[width=1\linewidth]{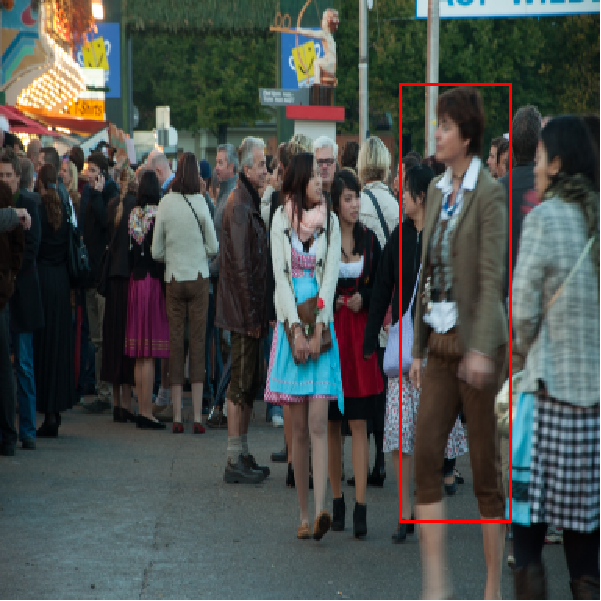}
\vspace{-0.75cm}
\caption*{Ground Truth}
\end{minipage}  
\begin{minipage}[b]{0.24\linewidth}
\centering
\includegraphics[width=1\linewidth]{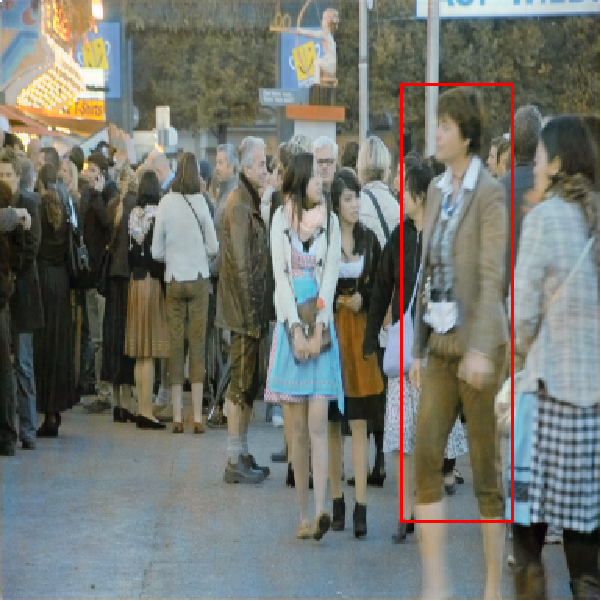}
\vspace{-0.75cm}
\caption*{FECNet}
\end{minipage}
\begin{minipage}[b]{0.24\linewidth}
\centering
\includegraphics[width=1\linewidth]{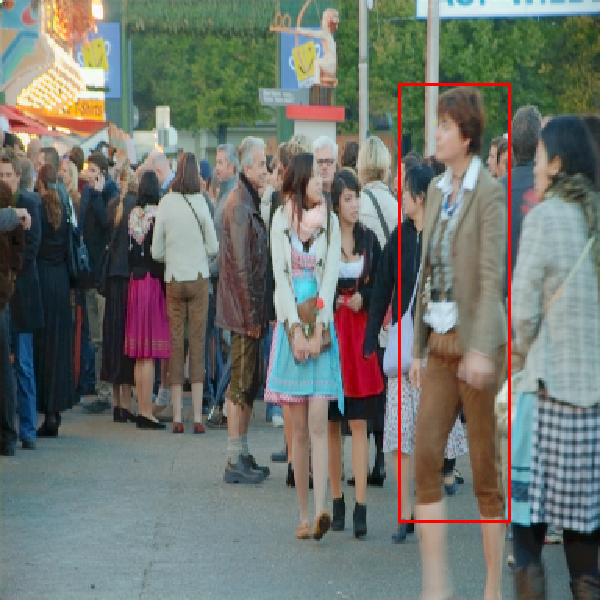}
\vspace{-0.75cm}
\caption*{FourLLIE}
\end{minipage} 
\begin{minipage}[b]{0.24\linewidth}
\centering
\includegraphics[width=1\linewidth]{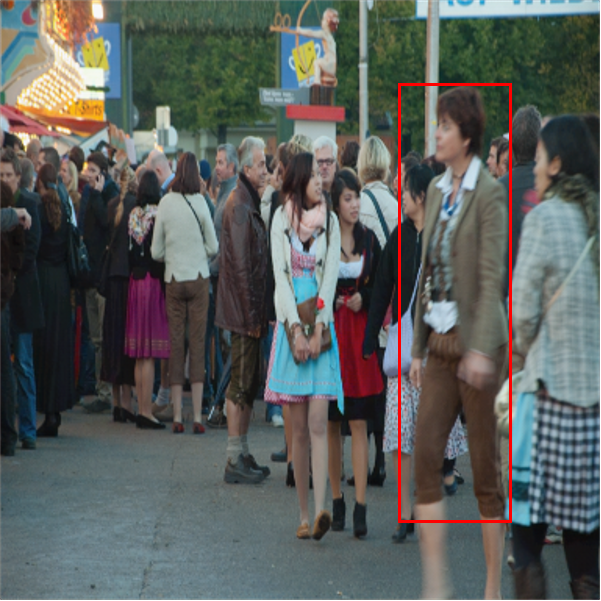}
\vspace{-0.75cm}
\caption*{DMFourLLIE}
\label{fourbasedcomparison-b}
\end{minipage}}
\vspace{-0.2cm}
\caption{Comparison with State-of-the-Art Two-Stage Fourier-Based Methods. (a) displays initial results from FECNet~\cite{four2} and FourLLIE~\cite{four1}, which exhibit color distortion and noise. Our method, in contrast, enhances brightness and preserves color fidelity, eliminating noise. (b) shows the superior noise suppression and color accuracy of our approach, particularly in highlighted areas.}
\vspace{-0.2cm}
\label{fig:com-one-stage}
\end{figure}

Due to environmental constraints and hardware limitations, images often suffer from issues like diminished visibility, uneven exposure, loss of detail, and color inaccuracies. These challenges significantly hinder the performance of computer vision applications, affecting critical tasks such as autonomous driving~\cite{sun2022shift}, pedestrian recognition~\cite{zhang2023diverse}, and object detection~\cite{hashmi2023featenhancer}. The primary aim of low-light image enhancement (LLIE) techniques is to restore images to a state of normal lighting, preserving as much texture detail as possible, thereby enhancing the efficiency of subsequent vision processing tasks. With the advancement of neural networks, traditional non-learning methods are becoming less prevalent. Recent years have seen the emergence of innovative LLIE methods \cite{lowlight1,lowlight2,lowlight3,lowlight4,lowlight5,lowlight6,lowlight7} in computer vision, yet a balance between method performance and computational demand remains elusive. High-performing models \cite{lowlight8,lowlight9} often come with substantial computational costs, whereas Fourier-based strategies \cite{four1,four2} present a cost-effective solution by leveraging Fourier frequency information for image brightening.

\begin{figure}
    \centering
    \includegraphics[width=1\linewidth]{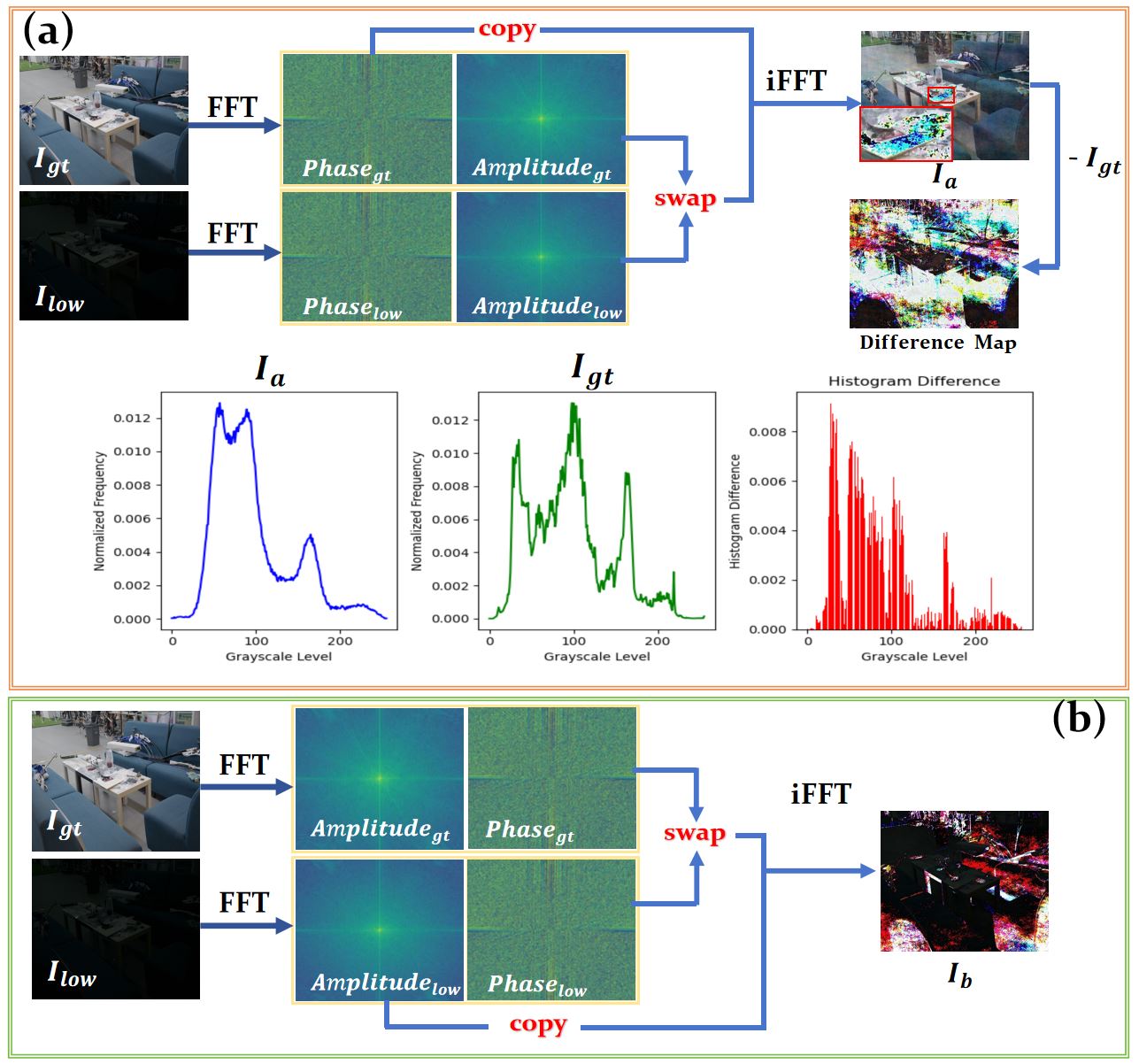}
    \caption{Our Motivations. (a): Illustrates issues from replacing the amplitude component of a low-light image with that of a normally illuminated image, resulting in $I_{a}$. (b): Highlights the importance of the phase component, as shown by the distorted image $I_{b}$ from replacing the phase component.}
    \label{fig:motivation}
    \vspace{-0.4cm}
\end{figure}

Researches have validated the efficacy of Fourier domain-based low-level image enhancement techniques across various tasks, including enhancement under low-light conditions \cite{four1,four2}, denoising \cite{fournoise1,fournoise2}, and super-resolution~\cite{foursuper}. In LLIE tasks, since luminance information is predominantly encoded in the amplitude component, common strategies \cite{four1,four2} involve amplifying the amplitude component and duplicating the phase component. However, the efficacy of such strategies warrants scrutiny. Through rigorous experimentation, we have arrived at the following two critical insights:

\textbf{(1) } As demonstrated in Fig. \ref{fig:motivation}(a), directly replacing the amplitude component of a low-light image with that of a normally illuminated image results in the generated image $I_{a}$ suffering from compromised brightness and texture, thereby introducing noise. Therefore, this indicates significant limitations in previous methods \cite{four1, four2}, where they directly approximate the amplitude component between the low-light input and the ground truth image. This approach leads to the degraded image $I_{a}$ being used as the learning target for the network, making the subsequent optimization challenging. We employ difference maps and histogram difference curves to visually demonstrate the disparity between $I_{a}$ and $I_{gt}$. Furthermore, enhancing the amplitude component may result in overflow of inherently brighter regions in the original image, introducing distorted patches in the visual output (highlighted by the red boxes in $I_{a}$). As consistently demonstrated in our results shown in Fig. \ref{fig:com-one-stage}, although FECNet \cite{four2} and FourLLIE \cite{four1} attempt to mitigate degradation by restoring the phase component and spatial information in the second stage, merely recovering these aspects does not fully address the distortions caused by the first stage. 

\textbf{(2) } As depicted in Fig. \ref{fig:motivation}(b), the distorted image $I_{b}$ demonstrates that the phase component also contains a significant amount of image information, which likewise requires meticulous handling during the process of frequency domain information transfer. The phase component retains more detailed information of the image, sometimes even more critical than the amplitude component.  Furthermore, swapping the phase components between two images while keeping their amplitude components unchanged results in the content of the images being exchanged along with the phase information, underscoring the significance of phase information for the structure and content of images. 

\begin{figure*}[ht]
    \centering
    \includegraphics[width=1\linewidth]{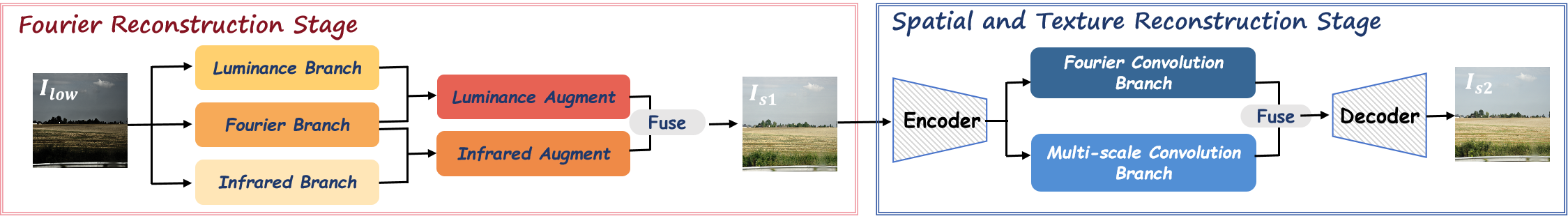}
    \vspace{-0.6cm}
    \caption{Overall framework of DMFourLLIE.}
    \label{fig:overall}
    \vspace{-0.35cm}
\end{figure*}

\begin{figure}
    \centering
    \includegraphics[width=1\linewidth]{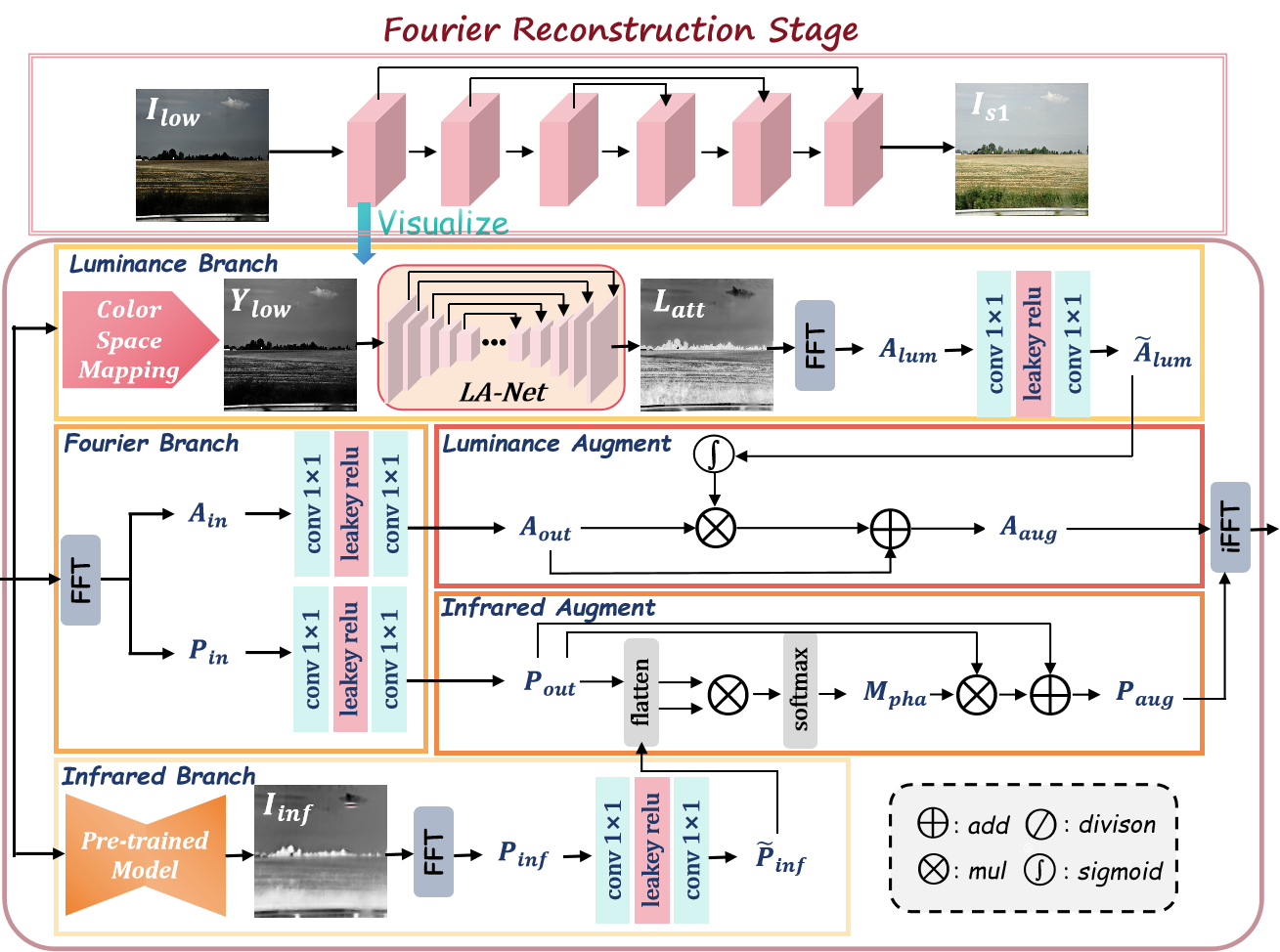}
    \vspace{-0.5cm}
    \caption{Structure of the Fourier Reconstruction Stage.}
    \label{fig:phase-one}
    \vspace{-0.4cm}
\end{figure}

\begin{figure}
    \centering
    \includegraphics[width=1\linewidth]{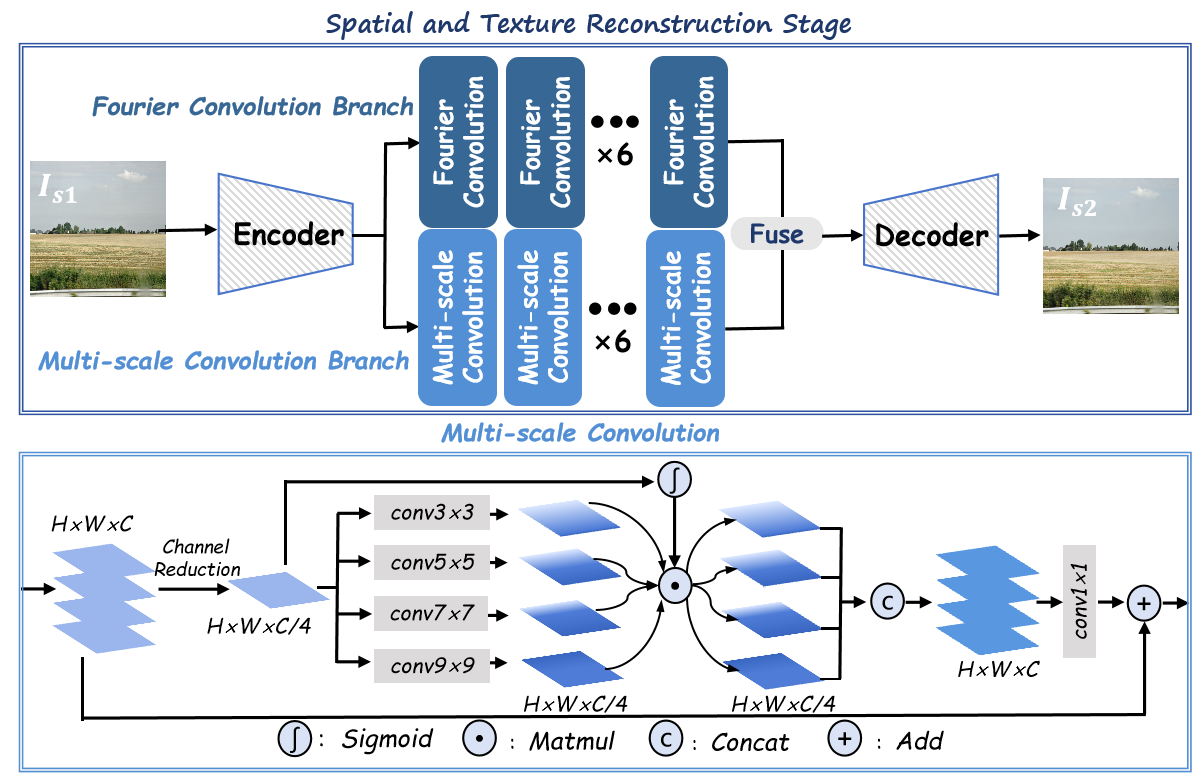}
    \vspace{-0.5cm}
    \caption{Structure of the Spatial and Texture Reconstruction Stage.}
    \label{fig:phase-two}
    \vspace{-0.4cm}
\end{figure}

Building upon the insights garnered, we propose a novel framework, DMFourLLIE (Dual-Stage Multi-Branch Fourier Low-Light Image Enhancement), which employs a two-stage design rooted in Fourier-based methods \cite{four1,four2,UHDFourICLR2023}, as illustrated in Fig. \ref{fig:overall}. The inaugural stage, detailed in Fig. \ref{fig:phase-one}, is dedicated to Fourier reconstruction. Within the Fourier frequency domain, our aim is to concurrently augment the expressiveness of both amplitude and phase components. To bolster the structural integrity of the phase component in low-light images, we integrate infrared prior information. For the amplitude component, we employ a luminance attention map to transition the luminance space, thereby enhancing the accuracy of amplification across various brightness levels.
Progressing to the second stage, depicted in Fig. \ref{fig:phase-two}, our focus shifts towards the reconstruction of spatial structures and textures. Here, we introduce a dual-pathway architecture that merges multi-scale spatial convolution branches with branches based on Fourier convolution \cite{chi2020fast}. This configuration is strategically designed to refine spatial structure representation while capturing intricate texture details.

Overall, the main contributions of this paper can be summarized as follows: 
\begin{itemize}
    \item In this paper, we introduce a framework named Dual-Stage Multi-Branch Fourier Low-Level Image Enhancement (DMFourLLIE). It addresses the limitations of existing methods by accurately enhancing frequency domain information through cross-modal enhancements within the Fourier domain. Additionally, it employs a dual-pathway architecture that combines multi-scale spatial convolution with Fourier convolution techniques to reconstruct fine-grained textures and spatial structures with high fidelity.  
    \item To the best of our knowledge, DMFourLLIE is the first approach in the LLIE field to leverage the synergistic potential of infrared and luminance priors in the Fourier space. By using cross-modal infrared images to guide the structural information of the phase component and luminance attention maps for precise amplitude component enhancement, our method introduces a novel strategy for embedding complementary perceptual priors within the Fourier space.  
    \item Extensive experiments validate the effectiveness of our hypotheses and the architectural advantages of DMFourLLIE. Our approach outperforms current state-of-the-art methods in the LLIE domain.
\end{itemize}

\section{Related Work}

\textbf{Non-learning and Learning-based LLIE Methods.} The spectrum of Low-Light Image Enhancement (LLIE) techniques is bifurcated into non-learning and learning-based methodologies. Traditional non-learning strategies encompass Histogram-based \cite{histogram1,histogram2,histogram3} and Retinex-based \cite{retinex1,retinex2,retinex3} methods. These approaches, however, fall short in accurately capturing scene dynamics and perceptual cues, leading to suboptimal brightness and color fidelity in real-world scenarios, along with exacerbated noise in extremely low-light conditions. The advent of deep learning catalyzed the development of a plethora of learning-based LLIE strategies \cite{lowlight1,lowlight2,lowlight3,lowlight4,lowlight5}. Notably, \cite{fu2023learning} introduced an unsupervised model that derives adaptive priors from pairs of low-light images. Similarly, \cite{wu2023learning} unveiled a semantic-aware framework that imbues models with a rich tapestry of semantic knowledge. Furthermore, studies \cite{four1,four2} have validated the efficacy of Fourier space manipulation for enhancing illumination in low-light scenes, achieving nuanced improvements with minimal model complexity.

\textbf{Infrared and Visible Image Fusion.} 
In computer vision, combining infrared and visible light imagery \cite{fuse1, fuse2} enhances visual representation in various tasks. However, the lack of annotated visible-infrared image pairs hinders the development of infrared-based deep learning models. This challenge has led to methods \cite{genfra1, genfra2} for training models in visible-to-infrared image translation. \cite{lee2023edge, Huang_2018_ECCV} introduced a translation model focused on edge preservation to maintain spatial textures. Unlike traditional shared encoder designs, \cite{zhao2023cddfuse} explored a novel approach for feature extraction and fusion. Inspired by the unique properties of infrared imagery and previous research, we investigate the application of infrared in the Fourier domain for illumination enhancement.

\textbf{Fourier-based LLIE Methods.} 
The exploration of Fourier frequency information for LLIE is still in its early stages. \cite{foursuper} investigated Fourier loss for super-resolution enhancement, while \cite{fournoise1} used Fourier transform for image fusion to combine spatial-frequency signals. \cite{fournoise2} tackled dehazing by exploring both frequency and spatial domains. Additionally, \cite{four2} highlighted the role of amplitude in encoding brightness and phase in capturing structural details, proposing a network for exposure correction. FourLLIE \cite{four1} showed that global feature extraction via Fourier analysis can be achieved without increasing model parameters. UHDFour \cite{UHDFourICLR2023} noted the similarity of amplitude components across image resolutions and proposed an illumination technique for ultra-high-definition content. Despite these advancements, the isolated manipulation of amplitude and phase components has limitations, highlighting the need for further exploration of Fourier frequency information.
\section{Method}     

\subsection{Fourier Frequency Information}
\label{sec:four}
Firstly, we provide a brief introduction to Fourier frequency information. Fourier frequency information refers to the representation of an input image $x$ in the Fourier space $X$ through a transform function $\mathcal{F}$. The input image $x$ has a shape of $H \times W$, where $H$ represents the height and $W$ represents the width. $\mathcal{F}$ is expressed as:
\begin{equation}
    \resizebox{1\hsize}{!}{$
        \mathcal{F}(x)(u,v)= X(u,v) = \frac{1}{\sqrt{H \times W}}\sum_{h=0}^{H-1}\sum_{w=0}^{W-1}x(h,w)e^{-j2\pi(\frac{h}{H}u+\frac{w}{W}v)}$,}
\end{equation}
where $h,w$ are the coordinates in the spatial space and $u,v$ are the coordinates in the Fourier space, $j$ is the imaginary unit, the inverse process $\mathcal{F}$ is denoted as $\mathcal{F}^{-1}$. Each complex component $X(u,v)$ can be represented by the amplitude component $\mathcal{A}(X(u,v))$ and phase component $\mathcal{P}(X(u,v))$. These two components are expressed as:
\begin{equation}
\mathcal{A}(X(u,v))=\sqrt{R^2(X(u,v))+I^2(X(u,v))},
\end{equation}
\begin{equation}
\mathcal{P}(X(u,v))=\arctan[\frac{I(X(u,v))}{R(X(u,v))}],
\end{equation}
where $R(x)$ and $I(x)$ represent the real and imaginary part of $X(u,v)$ respectively. In our method, the Fourier transform and inverse procedure is computed independently on each channel of feature maps.

\begin{figure*}
    \centering
    \includegraphics[width=1\linewidth]{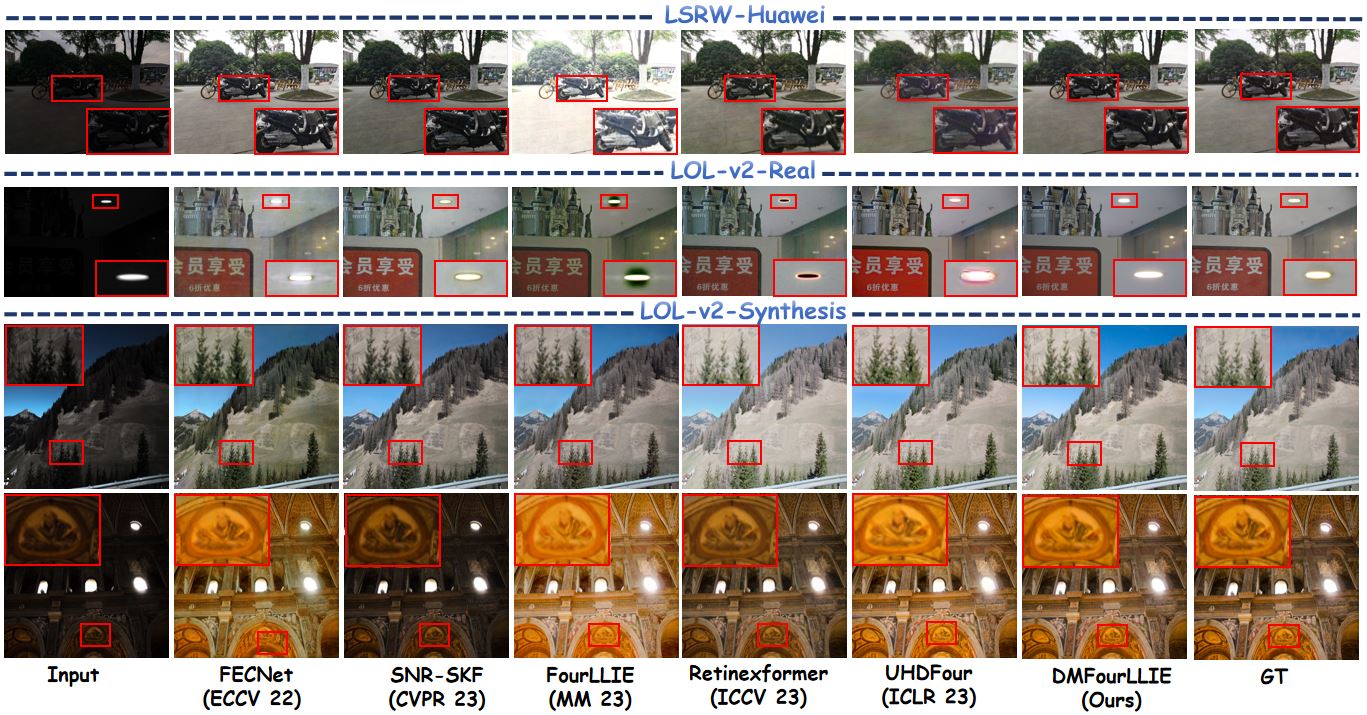}
    \vspace{-0.5cm}
    \caption{Visual comparison on LSRW-Huawei, LOL-v2-real, and LOL-v2-Synthesis datasets. DMFourLLIE effectively enhances visibility while preserving image texture and original colors without introducing additional noise.}
    \label{fig:com-v1}
\end{figure*}

\subsection{Fourier Reconstruction Stage}
As shown in Fig .\ref{fig:phase-one}, the low-light input image $I_{low}$ is fed into the Fourier reconstruction phase, resulting in the output of the first stage, $I_{s1}$. This stage consists of six identical components with skip connections, where the first component is visualized in detail. The LA-Net and pretrained models of Infrared Branch are only activated in the first component, with their outputs being passed and processed in the subsequent five components. The branches are described as follows:

\textbf{Fourier Branch:} The input image is first transformed into the Fourier space to obtain the amplitude component $A_{in}$ and the phase component $P_{in}$. Once transformed into the Fourier space, the assumption of spatial invariance no longer holds \cite{foursuper}. Therefore, we apply a $1 \times 1$ convolutional layer and LeakyReLU activation function in this space to extract the amplitude component, to avoid information loss. The amplitude and phase components are each extracted through a $1 \times 1$ convolutional layer with LeakyReLU activation, resulting in $A_{out}$ and $P_{out}$. Unlike existing methods \cite{four1,four2}, which only enhance the amplitude component and adopt a strategy of replicating the phase component, our analysis in Fig .\ref{fig:motivation} emphasizes the importance of phase information for the structure and content of images, indicating that merely replicating the phase component is not the best strategy.

\textbf{Luminance Branch:} To address the potential overflow issue in the Fourier domain space, we learn a luminance attention map through the luminance branch to precisely guide the enhancement level of the amplitude component in different areas, correctly enhancing underexposed areas while avoiding over-enhancement of correctly exposed areas. According to the study by Bread \cite{guo2023low}, among various color spaces, the luminance component (Y) in the YCbCr color space performs best and is least affected by noise. Inspired by this finding, we predict the luminance attention map $L_{att}$ through LA-Net (Luminance Attention Net) based on the luminance component $Y_{low}$ from the YCbCr color space. LA-Net follows the classic U-Net architecture with an encoder-decoder structure. This process and the learning target can be represented as:
\begin{equation}
L_{att}=LA\text{-}Net(Y_{low}),\quad\quad\quad
L_{att}=\frac{|Y_{low}-Y_{gt}|}{Y_{gt}},
\end{equation}
where $Y_{low}$ and $Y_{gt}$ represent the luminance components (Y channel) extracted from the input low-light image and the ground truth image, respectively. In the predicted $L_{att}$, brighter areas correspond to darker areas in the input low-light image, and vice versa. Then, $L_{att}$ is transformed into the Fourier space to obtain the amplitude component $A_{lum}$, and features are extracted through two $1 \times 1$ convolutional layers with LeakyReLU activation, resulting in $\tilde{A}_{lum}$.

\textbf{Luminance Augment:} To address the issue of varying brightness across different regions and the potential overflow problem during the transfer of amplitude components in the Fourier space, we propose a Luminance Augment operation. This operation leverages $L_{att}$ to make $A_{out}$ pay more attention to darker areas while reducing focus on brighter areas. Specifically, $\tilde{A}_{lum}$ undergoes a sigmoid operation and is then element-wise multiplied by $A{out}$, and through a residual connection, we obtain the enhanced amplitude component $A_{aug}$. The process can be described as follows: 
\begin{equation}
A_{aug}= A_{out}\cdot sigmoid(\tilde{A}_{lum}) + A_{out},
\end{equation}
through the aforementioned operation, $A_{aug}$ effectively adjusts the attention towards enhancing different areas, thereby more accurately amplifying the amplitude component.

\textbf{Infrared Branch:}
The input $I_{low}$ is processed by a pre-trained RGB-to-TIR image translation model \cite{lee2023edge, Huang_2018_ECCV}, which focuses on edge preservation. This model, trained on the VIPER\cite{Richter_2017_ICCV}, FLIR-ADAS, and STheReO\cite{9981857} datasets, generates infrared images from low-light scenes, improving infrared image performance in dark conditions. This process retains spatial texture information and generates the corresponding infrared image $I_{inf}$ from $I_{low}$. Infrared images, unaffected by illumination, color, and distortion, enhance object outlines. Since contour information primarily resides in the phase components of the Fourier domain, we extract the phase component $P_{inf}$ of $I_{inf}$ using FFT. Features are then extracted through two $1 \times 1$ convolutional layers with LeakyReLU activation, resulting in $\tilde{P}_{inf}$.

\textbf{Infrared Augment:} 
The phase component in the Fourier space is crucial for the structure and content of an image. Considering the rich structural information of infrared images and their high robustness in dark scene tasks, we are inspired by the transposed attention in Restormer~\cite{restormer} and design an Infrared Augment operation. This operation achieves phase attention maps $M_{pha}$ through cross-modal fusion of phase component features of infrared and visible light images in the Fourier space. The phase attention map $M_{pha}$ is described as follows:
\begin{equation}
    M_{pha}=softmax\left(flatten({P}_{out})\times flatten(\tilde{P}_{inf})\right),
\end{equation}
then, $P_{out}$ is multiplied by $M_{pha}$ and added through a residual connection, resulting in the enhanced phase component $P_{aug}$. This process is as follows:
\begin{equation}
    {P}_{aug}={P}_{out} \times M_{pha} + {P}_{out},
\end{equation}
to this end, we obtain the enhanced amplitude component $A_{aug}$ and phase component $P_{aug}$, which are then transformed back into the spatial domain through the inverse Fourier transform to yield the output $I_{s1}$:
\begin{equation}
    {I}_{s1} = {\mathcal F}^{-1}(A_{aug}\times\cos(P_{aug})+A_{aug}\times\sin(P_{aug})).%$} 
\end{equation}

\subsection{Spatial and Texture Reconstruction Stage}
While the Fourier space conveys global information, it lacks the enhancement of spatial details and may introduce additional noise and feature loss during the Fourier transform and feature transfer processes. To address this, we design the Spatial and Texture Reconstruction Stage to further refine image details and texture structures. Specifically, as shown in Fig. \ref{fig:phase-two}, this stage consists of two branches: the Multi-scale Convolution Branch and the Fourier Convolution Branch. The first-stage output $I_{s1}$ is initially downsampled by an encoder, then processed through dual-path branches composed of six Multi-scale Convolutions and Fourier Convolutions, respectively. Finally, features from both branches are concatenated and fed into a decoder, producing the final brightening result $I_{s2}$. These two branches work collaboratively to optimize pixel-level reconstruction, effectively recovering important details and structural patterns in the original image.

\textbf{Multi-scale Convolution Branch: }
The structure of the Multi-scale Convolution is depicted below Fig. \ref{fig:phase-two}. In this branch, multiple convolutional layers with varying kernel sizes are utilized to capture a wide range of spatial features across multiple scales and enhance the overall representation of spatial structures. To reduce computational cost, we downscale the dimension from (C=64) to (C/4=16), and gradually increase the kernel size by 2 from $3 \times 3$ to $9 \times 9$. After applying a sigmoid operation to the reduced features, they are multiplied with the output features of different convolutional kernels, then concatenated, restoring the dimension to (C=64). The features fused with multi-scale spatial information are processed through a $1 \times 1$ convolutional layer and combined with the input via a residual connection.

\textbf{Fourier Convolution Branch: }
Fast Fourier Convolution (FFC) \cite{chi2020fast} has been proven to handle high-resolution restoration cases with strong periodic texture, achieving robust enhancement at the resolution level. It consists of two branches: 1) a local branch that uses regular convolution, and 2) a global branch that convolves features after a fast Fourier transform. The outputs of both branches are then merged to obtain a larger receptive field and local invariance during the restoration process \cite{fourconvolution}. However, such a powerful model is unable to learn a reasonable overall spatial structure \cite{dong2022incremental}. Therefore, we pair it with the Multi-scale Convolution to form a dual-path structure, complementing it to achieve precise overall spatial structure through the final feature fusion.

\subsection{Loss Function}
To further improve image quality from both qualitative and quantitative aspects by considering perceptual information and regional differences, our loss function consists of four parts:
\begin{equation}
{\mathcal L}_{total}=\lambda_{1}{\mathcal L}_{s1}+\lambda_{2}{\mathcal L}_{s2} + \lambda_{3}{\mathcal L}_{per} + \lambda_{4}{\mathcal L}_{lum},
\end{equation}
where $\lambda$ denotes the loss weights, we empirically set $\lambda_{1},\lambda_{2},\lambda_{3},\lambda_{4}=[0.5,1,0.2,0.1]$. ${\mathcal L}_{s1} = \left\|I_{s1}-GT\right\|_2$ represents the $l_{2}$ loss between the preliminary enhanced result $I_{s1}$ and ground truth image $GT$. Similar to ${\mathcal L}_{s1}$, ${\mathcal L}_{s2} = \left\|I_{s2}-GT\right\|_2$ represents the loss between the final output $I_{s2}$ and $GT$.
${\mathcal L}_{per}$ is the perceptual loss between the $I_{s2}$ and $GT$, which constrains the features extracted from VGG~\cite{simonyan2014very} to obtain better visual results. 

Additionally, to acquire the luminance attention map for correctly guiding enhancement of luminance in image regions by $LA\text{-}Net$, we adopt $l_{2}$ error to measure the prediction error: 
\begin{equation}
{\mathcal L}_{lum} = \left\|LA\text{-}Net(I_{low})-L_{att}\right\|_2,
\end{equation}
where $LA\text{-}Net(I_{low})$ and $L_{att}$ represent the predicted and expected luminance attention maps, respectively.

\begin{table*}[ht]  
\centering  
\resizebox{\textwidth}{!}{  
\begin{tabular}{c|ccc|ccc|ccc|c|c}  
\toprule  
\multirow{2}{*}{\textbf{Methods}} & \multicolumn{3}{c|}{\textbf{LOL-v2-Real}}      & \multicolumn{3}{c|}{\textbf{LOL-v2-Syn}}      & \multicolumn{3}{c|}{\textbf{LSRW-Huawei}}   & \textbf{\#Param} & \textbf{\#FLOPs} \\ \cline{2-12}   
                                  & \multicolumn{1}{c}{PSNR ↑} & \multicolumn{1}{c}{SSIM ↑} & LPIPS ↓ & \multicolumn{1}{c}{PSNR ↑} & \multicolumn{1}{c}{SSIM ↑} & LPIPS ↓ & \multicolumn{1}{c}{PSNR ↑} & \multicolumn{1}{c}{SSIM ↑} & LPIPS ↓  & \textbf{(M)}   & \textbf{(G)}   \\ \midrule  
NPE~\cite{npe} \textcolor{orange}{(TIP, 13)}                  & 17.33   & 0.4642  & 0.2359  & 16.60   & 0.7781  & 0.1079  & 17.08   & 0.3905  & 0.2303  & -       & -       \\   
LIME~\cite{lime} \textcolor{orange}{(TIP, 16)}                & 15.24   & 0.4190  & 0.2203  & 16.88   & 0.7578  & 0.1041  & 17.00   & 0.3816  & 0.2069  & -       & -       \\   
SRIE~\cite{retinex1} \textcolor{orange}{(ICCV, 16)}           & 14.15   & 0.5524  & 0.2160  & 14.50   & 0.6640  & 0.1484  & 13.42   & 0.4282  & 0.2166  & -       & -       \\   
Kind~\cite{kind} \textcolor{orange}{(MM, 19)}                 & 20.01   & 0.8412  & 0.0813  & 22.62   & 0.9041  & 0.0515  & 16.58   & 0.5690  & 0.2259  & 8.02    & 34.99   \\   
MIRNet~\cite{lowlight9} \textcolor{orange}{(ECCV, 20)}        & 22.11   & 0.7942  & 0.1448  & 22.52   & 0.8997  & 0.0568  & 19.98   & 0.6085  & 0.2154  & 31.79   & 785.1   \\   
Kind++~\cite{kind++} \textcolor{orange}{(IJCV, 21)}           & 20.59   & 0.8294  & 0.0875  & 21.17   & 0.8814  & 0.0678  & 15.43   & 0.5695  & 0.2366  & 8.27    & -       \\   
SGM~\cite{lol} \textcolor{orange}{(TIP, 21)}                  & 20.06   & 0.8158  & 0.0727  & 22.05   & 0.9089  & 0.4841  & 18.85   & 0.5991  & 0.2492  & 2.31    & -       \\   
FECNet~\cite{four2} \textcolor{orange}{(ECCV, 22)}            & 20.67   & 0.7952  & 0.0995  & 22.57   & 0.8938  & 0.0699  & 21.09   & 0.6119  & 0.2341  & \textcolor{blue}{0.15}    & 5.82    \\   
HDMNet~\cite{liang2022learning} \textcolor{orange}{(MM, 22)}  & 18.55   & 0.7132  & 0.1717  & 20.54   & 0.8539  & 0.0690  & 20.81   & 0.6071  & 0.2375  & 2.32    & -       \\   
SNR-Aware~\cite{lowlight8} \textcolor{orange}{(CVPR, 22)}     & 21.48   & \textcolor{blue}{0.8478}  & 0.0740  & 24.13   & 0.9269  & 0.0318  & 20.67   & 0.5911  & 0.1923  & 39.12   & 26.35   \\   
Bread~\cite{guo2023low} \textcolor{orange}{(IJCV, 23)}        & 20.83   & 0.8217  & 0.0949  & 17.63   & 0.8376  & 0.0681  & 19.20   & 0.6179  & 0.2203  & 2.12    & \textcolor{red}{1.54}    \\   
FourLLIE~\cite{four1} \textcolor{orange}{(MM, 23)}            & 22.34   & 0.8403  & \textcolor{blue}{0.0573}  & 24.65   & 0.9192  & 0.0389  & 21.11   & 0.6256  & 0.1825  & \textcolor{red}{0.12}    & 4.07    \\   
SNR-SKF~\cite{wu2023learning} \textcolor{orange}{(CVPR, 23)}  & 20.66   & 0.8128  & 0.0757  & 17.21   & 0.7738  & 0.0731  & 16.21   & 0.5560  & \textcolor{red}{0.1615}  & 39.44   & 27.88   \\   
UHDFour~\cite{UHDFourICLR2023} \textcolor{orange}{(ICLR, 23)} & 19.42   & 0.7896  & 0.1151  & 23.64   & 0.8998  & 0.0341  & 19.39   & 0.6006  & 0.2466  & 17.54   & 4.78    \\   
Retinexformer~\cite{retinexformer} \textcolor{orange}{(ICCV, 23)} & \textcolor{red}{22.79} & 0.8397 & 0.0724 & \textcolor{blue}{25.67} & \textcolor{blue}{0.9295} & \textcolor{blue}{0.0273} & \textcolor{blue}{21.23} & \textcolor{blue}{0.6309} & \textcolor{blue}{0.1699} & 1.61 & 15.57 \\ \midrule  
DMFourLLIE \textcolor{orange}{(Ours)}                       & \textcolor{blue}{22.64} & \textcolor{red}{0.8589} & \textcolor{red}{0.0520} & \textcolor{red}{25.83} & \textcolor{red}{0.9314} & \textcolor{red}{0.0234} & \textcolor{red}{21.47} & \textcolor{red}{0.6331} & 0.1781 & 0.41 & \textcolor{blue}{1.69}   \\ \bottomrule  
\end{tabular}}  
\caption{Quantitative comparison on LOL-v2-Real~\cite{lol}, LOL-v2-Syn~\cite{lol}, and LSRW-Huawei~\cite{lsrw}. The best results are marked in red, the second-best are in blue.}  
\vspace{-0.5cm}  
\label{tab:com}  
\end{table*}  
    
\section{Experiments}
\subsection{Datasets and Experimental Setting}

\begin{table}[t]  
\centering  
\resizebox{0.9\linewidth}{!}{%  
\begin{tabular}{@{}lcccccc@{}}  
\toprule  
Methods       & LIME               & VV                 & DICM               & NPE                & MEF                & AVG                \\ \midrule  
Kind          & 4.772              & 3.835              & 3.614              & 4.175              & 4.819              & 4.194              \\ \midrule  
MIRNet        & 6.453              & 4.735              & 4.042              & 5.235              & 5.504              & 5.101              \\   
SGM           & 5.451              & 4.884              & 4.733              & 5.208              & 5.754              & 5.279              \\   
FECNet        & 6.041              & 3.346              & 4.139              & 4.500              & 4.707              & 4.336              \\   
HDMNet        & 6.403              & 4.462              & 4.773              & 5.108              & 5.993              & 5.056              \\   
Bread         & 4.717              & 3.304              & 4.179              & 4.160              & 5.369              & 4.194              \\   
Retinexformer & \textcolor{blue}{3.441} & 3.706              & 4.008              & \textcolor{blue}{3.893} & \textcolor{blue}{3.727} & \textcolor{blue}{3.755} \\   
FourLLIE      & 4.402              & \textcolor{red}{3.168} & \textcolor{red}{3.374} & 3.909              & 4.362              & 3.907              \\ \midrule  
DMFourLLIE    & \textcolor{red}{3.233} & \textcolor{blue}{3.298} & \textcolor{blue}{3.613} & \textcolor{red}{3.564} & \textcolor{red}{3.565} & \textcolor{red}{3.455} \\ \bottomrule  
\end{tabular}}  
\caption{NIQE scores on LIME, VV, DICM, NPE, and MEF datasets. The top results are highlighted in red and the second-best in blue. "AVG" denotes the average NIQE scores across these five datasets. All evaluated methods have been pre-trained on the LSRW-Huawei dataset.}  
\vspace{-0.4cm}  
\label{tab:comsmall}  
\end{table}  

DMFourLLIE is trained on three commonly used LLIE datasets: LOL-V2-Real~\cite{lol}, LOL-v2-synthesis~\cite{lol} and LSRW-Huawei~\cite{lsrw}. LOL-v2-Real is captured under varying exposure times and ISO in real scenes, including 689 low/normal light image pairs for training and 100 pairs for testing. LOL-v2-synthesis is synthesized from original images by analyzing the luminance channel distribution of low/normal light images. It contains 900 low light/normal light image pairs for training and 100 pairs for testing. Compared with the original LOL dataset~\cite{lolv1}, LOL-v2 is larger and more diverse, providing more convincing evaluation of performance. LSRW-Huawei was captured in real scenes using different devices, containing 3150 training image pairs and 20 test image pairs. Additionally, we also evaluated DMFourLLIE on five unpaired datasets: DICM~\cite{dicm} (64 images), LIME~\cite{lime} (10 images), MEF~\cite{mefl} (17 images), NPE~\cite{npe} (85 images) and VV(24 images).

\begin{table}[t]  
\centering  
\resizebox{0.95\linewidth}{!}{  
\begin{tabular}{@{}lccc@{}}  
\toprule  
\textbf{Settings}                    & \textbf{PSNR↑} & \textbf{SSIM↑} & \textbf{LPIPS↓}\\ \midrule  
w/o Fourier Reconstruction Stage             & 20.92     &  0.8435    &   \textbf{0.0473}    \\   
w/o Spatial and Texture Reconstruction Stage & 20.88     &  0.8337    &   0.0596    \\   
w/o Luminance Branch                         & 22.51     & 0.8504     &  0.0524    \\   
w/o Fourier Branch                           & 20.85     & 0.8391     &  0.0579    \\   
w/o Infrared Branch                          & 22.63     & 0.8476     &  0.0673    \\   
w/o Luminance Augment                        & 20.84     &  0.8382    &   0.0532    \\   
w/o Infrared Augment                         & 20.63     &  0.8317    &   0.0567    \\   
w/o Fourier Convolution Branch               & 22.46     & 0.8492     &  0.0576    \\   
w/o Multi-scale Convolution Branch           & 21.60     & 0.8404     &  0.0732    \\ \midrule  
DMFourLLIE                                   & \textbf{22.64} & \textbf{0.8585} & 0.0520 \\   
\bottomrule  
\end{tabular}}  
\caption{Ablation Studies on DMFourLLE with the LOL-v2-Real Dataset. The term 'w/o' denotes the absence of a specific component. 'w/o Fourier Branch' refers to a configuration that enhances only the amplitude component, similar to previous Fourier-based illumination networks, without involving the phase component. 'w/o Fourier Convolution Branch' and 'w/o Multi-scale Convolution Branch' represent scenarios where traditional convolutional layers replace these specialized modules.}  
\vspace{-0.4cm}  
\label{tab:ablationbranch}  
\end{table} 

DMFourLLIE adopts an end-to-end training strategy to jointly optimize the network parameters of both stages. We implement DMFourLLIE in Pytorch. During training, images are randomly cropped to $256 \times 256$ and augmented with random flipping. An ADAM optimizer with an initial learning rate of $4.0\times10^{-4}$ is used to optimize the network, along with a multi-step scheduler. The batch size is set to 8 and the total number of training iterations is set to $1.0\times10^{5}$. Training takes approximately 18 hours on an NVIDIA 3090 GPU.

\subsection{Comparison with Current Methods}
In this paper, our DMFourLLIE is compared to thirteen state-of-the-art LLIE methods, including traditional approaches LIME~\cite{lime}, NPE~\cite{npe} and SRIE~\cite{retinex1}, Fourier-based methods FourLLIE~\cite{four1}, FECNet~\cite{four2} and UHDFour~\cite{UHDFourICLR2023}. Deep learning-based methods involved in the comparison are Kind~\cite{kind}, Kind++~\cite{kind++}, MIRNet~\cite{lowlight9}, SGM~\cite{lol}, HDMNet~\cite{liang2022learning}, SNR-Aware~\cite{lowlight8}, Bread~\cite{guo2023low}, SNR-SKF~\cite{wu2023learning} and Retinexformer~\cite{retinexformer}. All deep learning models are trained on the same datasets using their original public codes, ensuring a fair evaluation.

\begin{figure}
    \centering
    \includegraphics[width=1\linewidth]{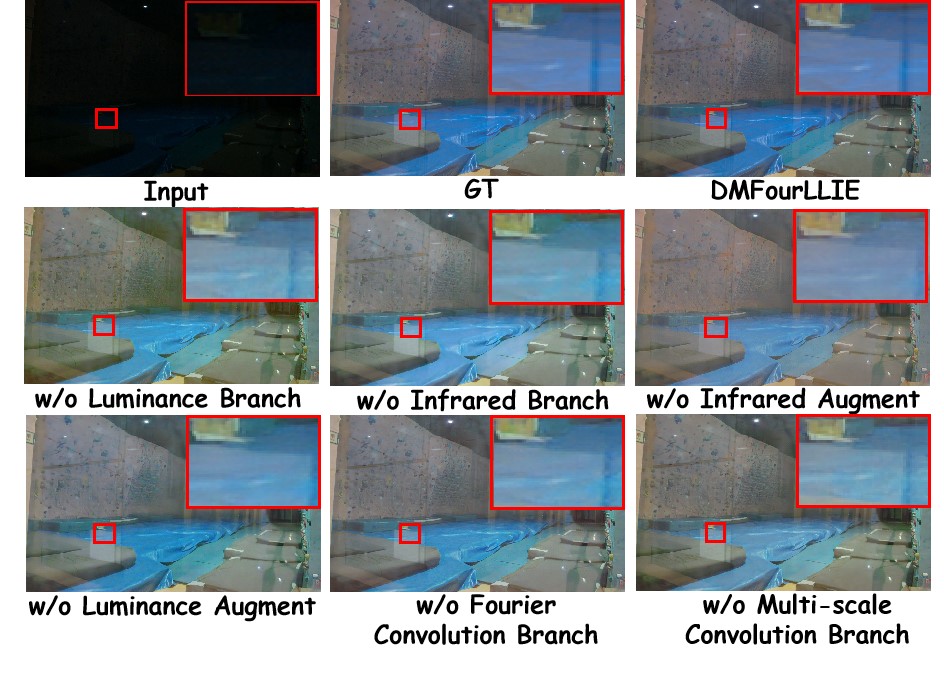}
    \vspace{-0.5cm}
    \caption{Visualizing Component Ablation in DMFourLLIE. Visual comparisons show noticeable differences against the ground truth when specific components of DMFourLLIE are removed. Each omitted component causes varying levels of color distortion and noise, highlighting their importance for high-fidelity image enhancement.}
    \label{fig:ab}
    \vspace{-0.3cm}
\end{figure}

\textbf{Quantitative results on LOL-v2-Real, LOL-v2-synthesis and LSRW-Huawei datasets.} In this work, we use Peak Signal-to-Noise Ratio (PSNR), Structural Similarity Index (SSIM)~\cite{SSIM} and Learned Perceptual Image Patch Similarity (LPIPS)~\cite{LP} as our evaluation metrics. Generally, higher PSNR and SSIM as well as lower LPIPS indicate higher similarity between two images.

As shown in Tab. \ref{tab:com}, we evaluate our approach using the LOL-v2-Real, LOL-v2-synthesis, and LSRW-Huawei datasets. Compared to the current state-of-the-art methods, it is evident that DMFourLLIE achieves nearly the best performance across the PSNR, SSIM, and LPIPS metrics. Furthermore, we also have advantages in terms of model parameters and computational efficiency.

\begin{table}[t]  
\centering  
\resizebox{0.95\linewidth}{!}{  
\begin{tabular}{@{}lcccccc@{}}  
\toprule  
\multirow{2}{*}{\textbf{Metrics}} & \multicolumn{6}{c}{\textbf{Loss Settings}} \\ \cmidrule(l){2-7}   
                         & w/o ${\mathcal L}_{s1}$ & w/o ${\mathcal L}_{s2}$ & w/o ${\mathcal L}_{lum}$ & w/o ${\mathcal L}_{per}$ & DMFourLLIE & $+{\mathcal L}_{amp}$ \\ \midrule  
PSNR↑                     & 21.61    & 20.41   & 21.98   & 21.31   & \textbf{22.64} & 22.33   \\   
SSIM↑                     & 0.8480   & 0.8419  & 0.8531  & 0.8319  & \textbf{0.8589} & 0.8403  \\   
\bottomrule  
\end{tabular}}  
\caption{Ablation experiments on the loss function.}  
\label{tab:loss}  
\vspace{-0.3cm}  
\end{table}  

\textbf{Visual Quality Comparisons.} Due to space limitations, we select the most recent and superior methods FECNet~\cite{four2}, SNR-SKF~\cite{wu2023learning}, FourLLIE~\cite{four1}, UHDFour~\cite{UHDFourICLR2023} and Retinexformer~\cite{retinexformer} for an intuitive comparison with our DMFourLLIE. Fig. \ref{fig:com-v1} shows the comparison results on LSRW-Huawei, LOL-v2-real, and LOL-v2-Synthesis datasets. In LSRW-Huawei, FourLLIE causes over-brightening leading to image overexposure, while SNR-SKF and Retinexformer do not achieve fully brightened visualization results. UHDFour exhibits an issue of insufficient brightness. FECNet, even with better brightening effects, lacks the clarity in motorcycle detail and overall image color (especially the floor) compared to our results, which are closer to ground truth (GT). In LOL-v2-real, FourLLIE and Retinexformer evidently introduce unavoidable blotchiness, greatly affecting the visualization results. FECNet and SNR-SKF show noticeable noise issues, whereas our method produces the clearest and most natural brightening results. In LOL-v2-synthesis, FECNet, SNR-SKF, FourLLIE, and Retinexformer suffer from insufficient brightness or overexposure issues, as well as some degree of color deviation. UHDFour introduces noise leading to blurry images, while our method achieves visually similar results to the GT.

\begin{table}[t]
\centering
\resizebox{0.85\linewidth}{!}{%
\begin{tabular}{@{}lccc@{}}
\toprule
        & FourLLIE & FourLLIE$_{s1}$ + DMFourLLIE$_{s2}$ & DMFourLLIE \\
\midrule
\bf{PSNR} ↑  & 24.65    & 24.94  (\textcolor{green}{↑ 0.29})   & \bf{25.83}      \\
\bf{SSIM} ↑  & 0.9192   & 0.9269 (\textcolor{green}{↑ 0.0077})   & \bf{0.9314}     \\
\bf{LPIPS} ↓ & 0.0389   & 0.0235 (\textcolor{green}{↓ 0.0154})   & \bf{0.0234}     \\
\bottomrule
\end{tabular}}
\caption{Objective validation of effectiveness by replacing the second stage of FourLLIE with our DMFourLLIE.}
\vspace{-0.3cm}  
\label{tab:comparison-s2}
\end{table}

\textbf{Quantitative results on LIME, VV, DICM, NPE and MEF datasets.} We measure the no-reference Natural Image Quality Evaluator (NIQE \cite{6353522}) scores on five non-paired datasets. Lower NIQE scores indicate images with higher naturalness quality. Tab. \ref{tab:comsmall} shows the NIQE evaluation results. It can be seen that our DMFourLLIE outperforms most of the existing LLIE methods.

\begin{figure}[t]
    \centering
    \includegraphics[width=1\linewidth]{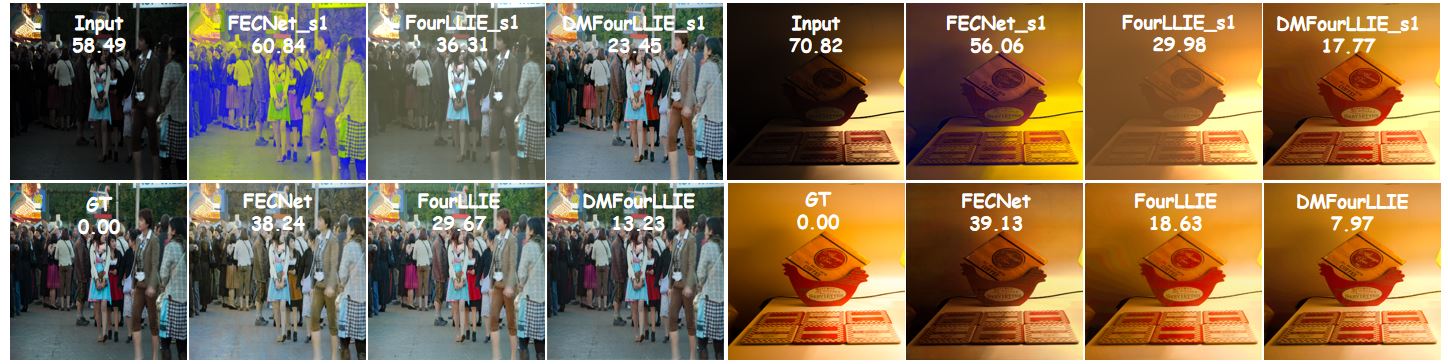}
    \caption{Stage-wise performance comparison.}
    \vspace{-0.3cm}  
    \label{fig:com_s1s2}
\end{figure}

\subsection{Effectiveness of the second stage}
\begin{figure}[t]
    \centering
    \includegraphics[width=1\linewidth]{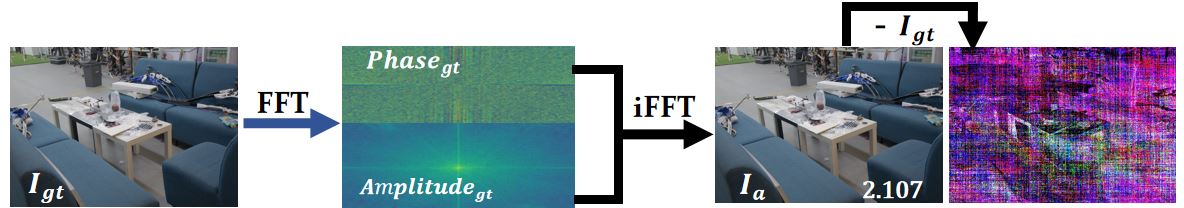}    
    \caption{Fourier transform loss. $I_{a}$ number is average error with GT.}
    \label{fig:com_dif}
    \vspace{-0.3cm}  
\end{figure}

We elaborate on the effectiveness of the second stage from three perspectives: (1) As shown qualitatively in Fig.\ref{fig:com_s1s2} and quantitatively through mean square error metrics, the second stage demonstrates its effectiveness and outperforms other methods; (2) By replacing the second stage of FourLLIE with the second stage of our DMFourLLIE, as reflected in Tab.\ref{tab:comparison-s2}, all performance metrics indicate improvements compared to the FourLLIE baseline; (3) Due to the inevitable information loss introduced during Fourier transform and its inverse in the first stage (refer to Fig.\ref{fig:com_dif}, owing to discretization), and after multiple transformations and network inferences, the refinement provided by the second stage is essential.

\subsection{Ablation Studies}
\textbf{Ablation experiments on components: }
In Tab. \ref{tab:ablationbranch} and Fig. \ref{fig:ab}, we conducted quantitative and qualitative ablation studies on each component of DMTourLLIE to assess their effectiveness. By controlling a single variable, we verified the effectiveness of each component and the overall network design.

\textbf{Ablation experiments on loss function: } 
To further analyze the impact of loss functions on model performance, we conducted an ablation study. As shown in Tab. \ref{tab:loss}, following the principle of controlling a single variable, we observed the contributions of different loss function configurations to model performance. The results indicate that among all compared configurations, DMFourLLIE achieved the best performance.

Moreover, to further validate the limitations in the design of the first-stage loss functions in FECNet and FourLLIE, we present the results of adding ${\mathcal L}_{amp}$ in the last column of Tab. \ref{tab:loss} (following the design of FECNet and FourLLIE, where ${\mathcal L}_{amp}$ is defined as the difference between the enhanced amplitude component $A_{aug}$ and the ground truth amplitude component $A_{gt}$). We found that the performance actually decreases after adding ${\mathcal L}_{amp}$, which again supports our motivational hypothesis that the strategy of solely boosting the amplitude component leads to using degraded images as the learning target for the network.

\subsection{Impact of different pre-trained model}
 We examined the effects of different models on infrared information extraction, and the impact of incorporating depth and edge maps. In Tab.\ref{tab:com_map}, we provide a comprehensive analysis of the impact of different models on infrared information extraction, including the use of depth maps and edge maps.
 It is well-known that infrared images have significant advantages in extremely low-light conditions. In contrast, depth maps and edge maps are often susceptible to interference from factors such as lighting and color noise. This validates our motivation for using infrared images to enhance the structural integrity of the phase component. The infrared extraction model employed by DMFourLLIE achieves optimal overall performance.

 \begin{table}[t]
\centering
\resizebox{\linewidth}{!}{%
\begin{tabular}{@{}lccccc@{}}
\toprule
        & \begin{tabular}[c]{@{}c@{}}Infrared Image\\ (StawGAN \cite{stawgan})\end{tabular} & \begin{tabular}[c]{@{}c@{}}Infrared Image\\ (PearlGAN \cite{luo2022thermal})\end{tabular} & Depth Map \cite{depthanything} & Edge Map \cite{xu2023low} & DMFourLLIE \\ \midrule
\bf{PSNR} ↑  & 25.17                                                               & \textcolor{blue}{25.32}                                                                & 24.97     & 25.15    & \textcolor{red}{25.83}      \\
\bf{SSIM} ↑  & 0.9287                                                              & \textcolor{blue}{0.9307}                                                              & 0.9256    & 0.9275   & \textcolor{red}{0.9317}     \\
\bf{LPIPS} ↓ & 0.0265                                                              & \textcolor{red}{0.0225}                                                               & 0.0238    & 0.0245   & \textcolor{blue}{0.0234}     \\ \bottomrule
\end{tabular}
}
\caption{The impact of infrared image , depth map , and edge map on model performance was investigated. Comparative analysis revealed that infrared images achieve superior performance. }
\label{tab:com_map}
\vspace{-0.3cm}
\end{table}

\begin{table}[t]  
\centering  
\resizebox{0.7\linewidth}{!}{  
\begin{tabular}{@{}lccc@{}}  
\toprule  
Metrics & FourLLIE & UHDFour & DMFourLLIE \\ \midrule  
UICM ↑  & 0.9133   & 0.7464  & \textbf{0.9426} \\   
NIQE ↓  & 3.143    & 5.385   & \textbf{3.047}  \\   
\bottomrule  
\end{tabular}}  
\caption{Visualization quality comparison on Dark Face.}  
\label{tab:uicm-niqe}  
\vspace{-0.3cm}  
\end{table}

\subsection{Object Detection and Visualization}

\begin{figure}[t]
    \centering
    \includegraphics[width=1\linewidth]{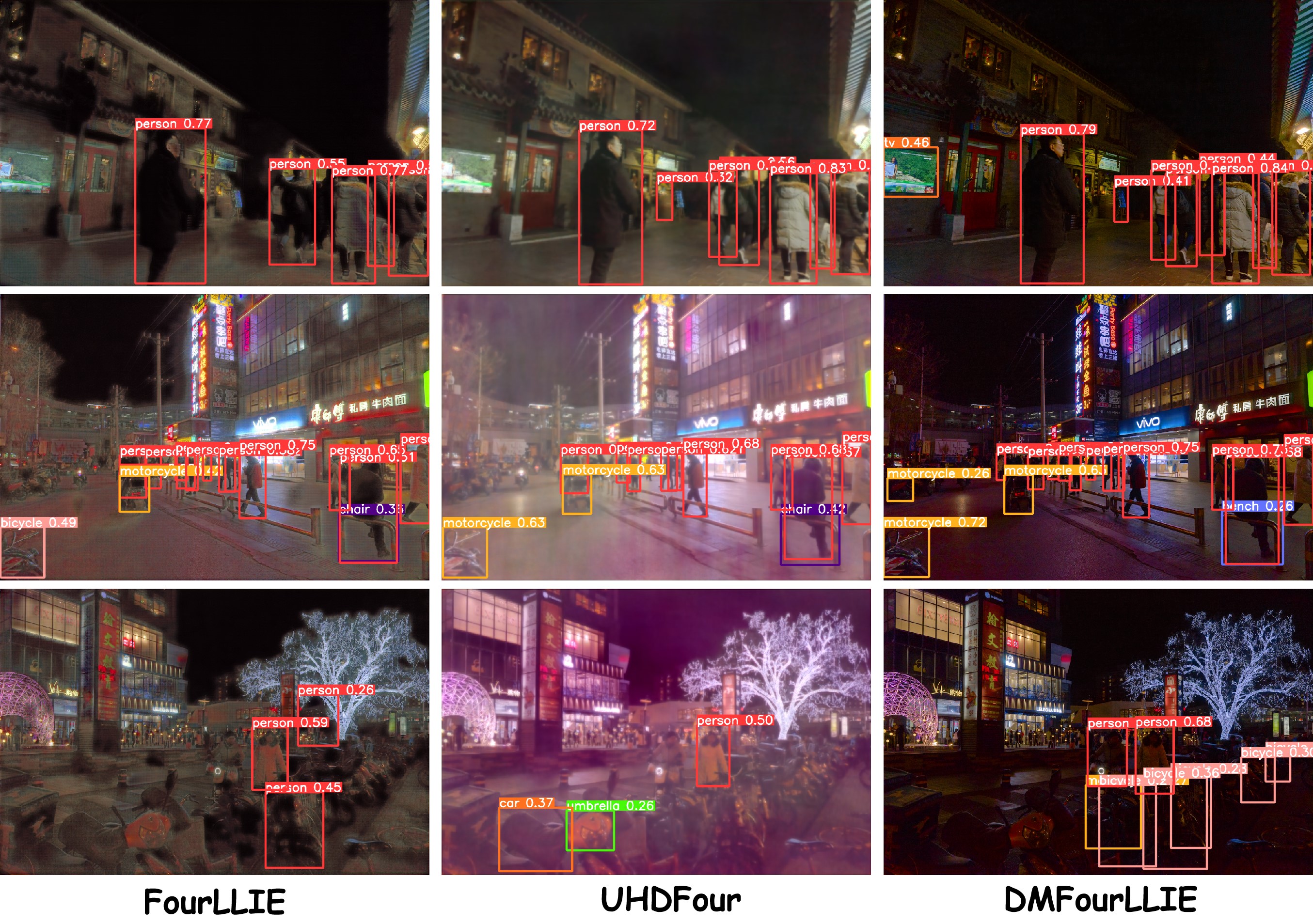}
    \caption{Detection comparison results on the Dark Face dataset.}
    \vspace{-0.3cm}
    \label{fig:detection-vis}
\end{figure}

On the Dark Face dataset ~\cite{yang2020advancing}, we conducted object detection experiments under low-light conditions to evaluate the preprocessing effects of the latest Fourier-based methods on advanced visual understanding tasks. The Dark Face dataset consists of 6000 low-light images with real-world annotations. For testing, we randomly selected 200 images from the dataset and validated using the official YOLOv5 model pre-trained on the COCO~\cite{lin2014microsoft} dataset. Fig. \ref{fig:detection-vis} presents the visual comparison results, where our method outperforms other methods significantly in terms of detection accuracy and recall rate. Notably, in the results of the second row, DMFourLLIE successfully detects the backpack, while in the fourth row, the detection rate of bicycles is significantly higher than other methods, with other methods showing instances of false detections.

Furthermore, through visual analysis, it is evident that DMFourLLIE surpasses other methods in terms of brightness enhancement, naturalness, and clarity. This is particularly noticeable when observing the faces in the second and fifth rows. In contrast, both FourLLIE and UHDFour methods exhibit blurriness and noise issues. Therefore, as shown in Tab. \ref{tab:uicm-niqe}, we also compared image quality metrics (NIQE \cite{6353522} and UICM \cite{7305804}) on the Dark Face dataset to further support the advantages of our method.
\section{Conclusion}
In this paper, we introduce the Dual-Stage Multi-Branch Fourier Low-Light Image Enhancement (DMFourLLIE) framework, a novel structure designed to enhance the expressiveness of Fourier frequency domain information and improve the spatial and textural details of images. The first stage of our framework utilizes a distinctive multi-branch structure that integrates infrared and brightness priors, thereby refining the expressiveness and accuracy of frequency domain information. This integration significantly bolsters the phase component's capacity to preserve image structure while enabling precise amplification of the amplitude component, tailored to regional luminance variations. In the second stage, DMFourLLIE employs a multi-scale spatial perception module in conjunction with the innovative application of fast Fourier convolution. This dual-branch approach markedly enhances the representation of spatial structures and the delineation of subtle texture details in the enhanced images, leading to superior image quality.

\begin{acks}
This work was supported by National Natural Science Foundation of China (62071199) and Jilin Province Industrial Key Core Technology Tackling Project (20230201085GX).
\end{acks}


\bibliographystyle{ACM-Reference-Format}
\bibliography{reference}


\begin{thebibliography}{62}

%%% the Web address.

\ifx \showCODEN    \undefined \def \showCODEN     #1{\unskip}     \fi
\ifx \showDOI      \undefined \def \showDOI       #1{#1}\fi
\ifx \showISBNx    \undefined \def \showISBNx     #1{\unskip}     \fi
\ifx \showISBNxiii \undefined \def \showISBNxiii  #1{\unskip}     \fi
\ifx \showISSN     \undefined \def \showISSN      #1{\unskip}     \fi
\ifx \showLCCN     \undefined \def \showLCCN      #1{\unskip}     \fi
\ifx \shownote     \undefined \def \shownote      #1{#1}          \fi
\ifx \showarticletitle \undefined \def \showarticletitle #1{#1}   \fi
\ifx \showURL      \undefined \def \showURL       {\relax}        \fi
% invisible to TeX
\providecommand\bibfield[2]{#2}
\providecommand\bibinfo[2]{#2}
\providecommand\natexlab[1]{#1}
\providecommand\showeprint[2][]{arXiv:#2}

\bibitem[Arici et~al\mbox{.}(2009)]%
        {histogram1}
\bibfield{author}{\bibinfo{person}{Tarik Arici}, \bibinfo{person}{Salih Dikbas}, {and} \bibinfo{person}{Yucel Altunbasak}.} \bibinfo{year}{2009}\natexlab{}.
\newblock \showarticletitle{A histogram modification framework and its application for image contrast enhancement}.
\newblock \bibinfo{journal}{\emph{IEEE TIP}} \bibinfo{volume}{18}, \bibinfo{number}{9} (\bibinfo{year}{2009}), \bibinfo{pages}{1921--1935}.
\newblock


\bibitem[Cai et~al\mbox{.}(2023)]%
        {retinexformer}
\bibfield{author}{\bibinfo{person}{Yuanhao Cai}, \bibinfo{person}{Hao Bian}, \bibinfo{person}{Jing Lin}, \bibinfo{person}{Haoqian Wang}, \bibinfo{person}{Radu Timofte}, {and} \bibinfo{person}{Yulun Zhang}.} \bibinfo{year}{2023}\natexlab{}.
\newblock \showarticletitle{Retinexformer: One-stage Retinex-based Transformer for Low-light Image Enhancement}. In \bibinfo{booktitle}{\emph{ICCV}}.
\newblock


\bibitem[Cao et~al\mbox{.}(2023)]%
        {fuse1}
\bibfield{author}{\bibinfo{person}{Bing Cao}, \bibinfo{person}{Yiming Sun}, \bibinfo{person}{Pengfei Zhu}, {and} \bibinfo{person}{Qinghua Hu}.} \bibinfo{year}{2023}\natexlab{}.
\newblock \showarticletitle{Multi-Modal Gated Mixture of Local-to-Global Experts for Dynamic Image Fusion}. In \bibinfo{booktitle}{\emph{ICCV}}. \bibinfo{pages}{23555--23564}.
\newblock


\bibitem[Chi et~al\mbox{.}(2020)]%
        {chi2020fast}
\bibfield{author}{\bibinfo{person}{Lu Chi}, \bibinfo{person}{Borui Jiang}, {and} \bibinfo{person}{Yadong Mu}.} \bibinfo{year}{2020}\natexlab{}.
\newblock \showarticletitle{Fast fourier convolution}.
\newblock \bibinfo{journal}{\emph{Advances in Neural Information Processing Systems}}  \bibinfo{volume}{33} (\bibinfo{year}{2020}), \bibinfo{pages}{4479--4488}.
\newblock


\bibitem[Dong et~al\mbox{.}(2022)]%
        {dong2022incremental}
\bibfield{author}{\bibinfo{person}{Qiaole Dong}, \bibinfo{person}{Chenjie Cao}, {and} \bibinfo{person}{Yanwei Fu}.} \bibinfo{year}{2022}\natexlab{}.
\newblock \showarticletitle{Incremental transformer structure enhanced image inpainting with masking positional encoding}. In \bibinfo{booktitle}{\emph{CVPR}}. \bibinfo{pages}{11358--11368}.
\newblock


\bibitem[Fu et~al\mbox{.}(2016)]%
        {retinex1}
\bibfield{author}{\bibinfo{person}{Xueyang Fu}, \bibinfo{person}{Delu Zeng}, \bibinfo{person}{Yue Huang}, \bibinfo{person}{Xiao-Ping Zhang}, {and} \bibinfo{person}{Xinghao Ding}.} \bibinfo{year}{2016}\natexlab{}.
\newblock \showarticletitle{A weighted variational model for simultaneous reflectance and illumination estimation}. In \bibinfo{booktitle}{\emph{CVPR}}. \bibinfo{pages}{2782--2790}.
\newblock


\bibitem[Fu et~al\mbox{.}(2023)]%
        {fu2023learning}
\bibfield{author}{\bibinfo{person}{Zhenqi Fu}, \bibinfo{person}{Yan Yang}, \bibinfo{person}{Xiaotong Tu}, \bibinfo{person}{Yue Huang}, \bibinfo{person}{Xinghao Ding}, {and} \bibinfo{person}{Kai-Kuang Ma}.} \bibinfo{year}{2023}\natexlab{}.
\newblock \showarticletitle{Learning a Simple Low-Light Image Enhancer From Paired Low-Light Instances}. In \bibinfo{booktitle}{\emph{CVPR}}. \bibinfo{pages}{22252--22261}.
\newblock


\bibitem[Fuoli et~al\mbox{.}(2021)]%
        {foursuper}
\bibfield{author}{\bibinfo{person}{Dario Fuoli}, \bibinfo{person}{Luc Van~Gool}, {and} \bibinfo{person}{Radu Timofte}.} \bibinfo{year}{2021}\natexlab{}.
\newblock \showarticletitle{Fourier space losses for efficient perceptual image super-resolution}. In \bibinfo{booktitle}{\emph{ICCV}}. \bibinfo{pages}{2360--2369}.
\newblock


\bibitem[Guo et~al\mbox{.}(2020)]%
        {lowlight1}
\bibfield{author}{\bibinfo{person}{Chunle Guo}, \bibinfo{person}{Chongyi Li}, \bibinfo{person}{Jichang Guo}, \bibinfo{person}{Chen~Change Loy}, \bibinfo{person}{Junhui Hou}, \bibinfo{person}{Sam Kwong}, {and} \bibinfo{person}{Runmin Cong}.} \bibinfo{year}{2020}\natexlab{}.
\newblock \showarticletitle{Zero-reference deep curve estimation for low-light image enhancement}. In \bibinfo{booktitle}{\emph{CVPR}}. \bibinfo{pages}{1780--1789}.
\newblock


\bibitem[Guo and Hu(2023)]%
        {guo2023low}
\bibfield{author}{\bibinfo{person}{Xiaojie Guo} {and} \bibinfo{person}{Qiming Hu}.} \bibinfo{year}{2023}\natexlab{}.
\newblock \showarticletitle{Low-light image enhancement via breaking down the darkness}.
\newblock \bibinfo{journal}{\emph{IJCV}} \bibinfo{volume}{131}, \bibinfo{number}{1} (\bibinfo{year}{2023}), \bibinfo{pages}{48--66}.
\newblock


\bibitem[Guo et~al\mbox{.}(2016)]%
        {lime}
\bibfield{author}{\bibinfo{person}{Xiaojie Guo}, \bibinfo{person}{Yu Li}, {and} \bibinfo{person}{Haibin Ling}.} \bibinfo{year}{2016}\natexlab{}.
\newblock \showarticletitle{LIME: Low-light image enhancement via illumination map estimation}.
\newblock \bibinfo{journal}{\emph{IEEE TIP}} \bibinfo{volume}{26}, \bibinfo{number}{2} (\bibinfo{year}{2016}), \bibinfo{pages}{982--993}.
\newblock


\bibitem[Hai et~al\mbox{.}(2023)]%
        {lsrw}
\bibfield{author}{\bibinfo{person}{Jiang Hai}, \bibinfo{person}{Zhu Xuan}, \bibinfo{person}{Ren Yang}, \bibinfo{person}{Yutong Hao}, \bibinfo{person}{Fengzhu Zou}, \bibinfo{person}{Fang Lin}, {and} \bibinfo{person}{Songchen Han}.} \bibinfo{year}{2023}\natexlab{}.
\newblock \showarticletitle{R2rnet: Low-light image enhancement via real-low to real-normal network}.
\newblock \bibinfo{journal}{\emph{Journal of Visual Communication and Image Representation}}  \bibinfo{volume}{90} (\bibinfo{year}{2023}), \bibinfo{pages}{103712}.
\newblock


\bibitem[Hashmi et~al\mbox{.}(2023)]%
        {hashmi2023featenhancer}
\bibfield{author}{\bibinfo{person}{Khurram~Azeem Hashmi}, \bibinfo{person}{Goutham Kallempudi}, \bibinfo{person}{Didier Stricker}, {and} \bibinfo{person}{Muhammad~Zeshan Afzal}.} \bibinfo{year}{2023}\natexlab{}.
\newblock \showarticletitle{FeatEnHancer: Enhancing Hierarchical Features for Object Detection and Beyond Under Low-Light Vision}. In \bibinfo{booktitle}{\emph{CVPR}}. \bibinfo{pages}{6725--6735}.
\newblock


\bibitem[Huang et~al\mbox{.}(2022)]%
        {four2}
\bibfield{author}{\bibinfo{person}{Jie Huang}, \bibinfo{person}{Yajing Liu}, \bibinfo{person}{Feng Zhao}, \bibinfo{person}{Keyu Yan}, \bibinfo{person}{Jinghao Zhang}, \bibinfo{person}{Yukun Huang}, \bibinfo{person}{Man Zhou}, {and} \bibinfo{person}{Zhiwei Xiong}.} \bibinfo{year}{2022}\natexlab{}.
\newblock \showarticletitle{Deep fourier-based exposure correction network with spatial-frequency interaction}. In \bibinfo{booktitle}{\emph{ECCV}}. Springer, \bibinfo{pages}{163--180}.
\newblock


\bibitem[Huang et~al\mbox{.}(2018)]%
        {Huang_2018_ECCV}
\bibfield{author}{\bibinfo{person}{Xun Huang}, \bibinfo{person}{Ming-Yu Liu}, \bibinfo{person}{Serge Belongie}, {and} \bibinfo{person}{Jan Kautz}.} \bibinfo{year}{2018}\natexlab{}.
\newblock \showarticletitle{Multimodal Unsupervised Image-to-image Translation}. In \bibinfo{booktitle}{\emph{ECCV}}.
\newblock


\bibitem[Kniaz et~al\mbox{.}(2018)]%
        {genfra2}
\bibfield{author}{\bibinfo{person}{Vladimir~V Kniaz}, \bibinfo{person}{Vladimir~A Knyaz}, \bibinfo{person}{Jiri Hladuvka}, \bibinfo{person}{Walter~G Kropatsch}, {and} \bibinfo{person}{Vladimir Mizginov}.} \bibinfo{year}{2018}\natexlab{}.
\newblock \showarticletitle{Thermalgan: Multimodal color-to-thermal image translation for person re-identification in multispectral dataset}. In \bibinfo{booktitle}{\emph{ECCV}}. \bibinfo{pages}{0--0}.
\newblock


\bibitem[Lee et~al\mbox{.}(2012)]%
        {dicm}
\bibfield{author}{\bibinfo{person}{Chulwoo Lee}, \bibinfo{person}{Chul Lee}, {and} \bibinfo{person}{Chang-Su Kim}.} \bibinfo{year}{2012}\natexlab{}.
\newblock \showarticletitle{Contrast enhancement based on layered difference representation}. In \bibinfo{booktitle}{\emph{ICIP}}. IEEE, \bibinfo{pages}{965--968}.
\newblock


\bibitem[Lee et~al\mbox{.}(2013)]%
        {histogram2}
\bibfield{author}{\bibinfo{person}{Chulwoo Lee}, \bibinfo{person}{Chul Lee}, {and} \bibinfo{person}{Chang-Su Kim}.} \bibinfo{year}{2013}\natexlab{}.
\newblock \showarticletitle{Contrast enhancement based on layered difference representation of 2D histograms}.
\newblock \bibinfo{journal}{\emph{IEEE TIP}} \bibinfo{volume}{22}, \bibinfo{number}{12} (\bibinfo{year}{2013}), \bibinfo{pages}{5372--5384}.
\newblock


\bibitem[Lee et~al\mbox{.}(2023)]%
        {lee2023edge}
\bibfield{author}{\bibinfo{person}{Dong-Guw Lee}, \bibinfo{person}{Myung-Hwan Jeon}, \bibinfo{person}{Younggun Cho}, {and} \bibinfo{person}{Ayoung Kim}.} \bibinfo{year}{2023}\natexlab{}.
\newblock \showarticletitle{Edge-guided multi-domain rgb-to-tir image translation for training vision tasks with challenging labels}. In \bibinfo{booktitle}{\emph{2023 IEEE International Conference on Robotics and Automation (ICRA)}}. IEEE, \bibinfo{pages}{8291--8298}.
\newblock


\bibitem[Li et~al\mbox{.}(2023)]%
        {UHDFourICLR2023}
\bibfield{author}{\bibinfo{person}{Chongyi Li}, \bibinfo{person}{Chun-Le Guo}, \bibinfo{person}{Man Zhou}, \bibinfo{person}{Zhexin Liang}, \bibinfo{person}{Shangchen Zhou}, \bibinfo{person}{Ruicheng Feng}, {and} \bibinfo{person}{Chen~Change Loy}.} \bibinfo{year}{2023}\natexlab{}.
\newblock \showarticletitle{EmbeddingFourier for Ultra-High-Definition Low-Light Image Enhancement}. In \bibinfo{booktitle}{\emph{ICLR}}.
\newblock


\bibitem[Li et~al\mbox{.}(2018)]%
        {retinex2}
\bibfield{author}{\bibinfo{person}{Mading Li}, \bibinfo{person}{Jiaying Liu}, \bibinfo{person}{Wenhan Yang}, \bibinfo{person}{Xiaoyan Sun}, {and} \bibinfo{person}{Zongming Guo}.} \bibinfo{year}{2018}\natexlab{}.
\newblock \showarticletitle{Structure-revealing low-light image enhancement via robust retinex model}.
\newblock \bibinfo{journal}{\emph{IEEE TIP}} \bibinfo{volume}{27}, \bibinfo{number}{6} (\bibinfo{year}{2018}), \bibinfo{pages}{2828--2841}.
\newblock


\bibitem[Liang et~al\mbox{.}(2022)]%
        {liang2022learning}
\bibfield{author}{\bibinfo{person}{Yudong Liang}, \bibinfo{person}{Bin Wang}, \bibinfo{person}{Wenqi Ren}, \bibinfo{person}{Jiaying Liu}, \bibinfo{person}{Wenjian Wang}, {and} \bibinfo{person}{Wangmeng Zuo}.} \bibinfo{year}{2022}\natexlab{}.
\newblock \showarticletitle{Learning hierarchical dynamics with spatial adjacency for image enhancement}. In \bibinfo{booktitle}{\emph{ACM MM}}. \bibinfo{pages}{2767--2776}.
\newblock


\bibitem[Lin et~al\mbox{.}(2014)]%
        {lin2014microsoft}
\bibfield{author}{\bibinfo{person}{Tsung-Yi Lin}, \bibinfo{person}{Michael Maire}, \bibinfo{person}{Serge Belongie}, \bibinfo{person}{James Hays}, \bibinfo{person}{Pietro Perona}, \bibinfo{person}{Deva Ramanan}, \bibinfo{person}{Piotr Doll{\'a}r}, {and} \bibinfo{person}{C~Lawrence Zitnick}.} \bibinfo{year}{2014}\natexlab{}.
\newblock \showarticletitle{Microsoft coco: Common objects in context}. In \bibinfo{booktitle}{\emph{ECCV}}. Springer, \bibinfo{pages}{740--755}.
\newblock


\bibitem[Luo et~al\mbox{.}(2022)]%
        {luo2022thermal}
\bibfield{author}{\bibinfo{person}{Fuya Luo}, \bibinfo{person}{Yunhan Li}, \bibinfo{person}{Guang Zeng}, \bibinfo{person}{Peng Peng}, \bibinfo{person}{Gang Wang}, {and} \bibinfo{person}{Yongjie Li}.} \bibinfo{year}{2022}\natexlab{}.
\newblock \showarticletitle{Thermal infrared image colorization for nighttime driving scenes with top-down guided attention}.
\newblock \bibinfo{journal}{\emph{IEEE Transactions on Intelligent Transportation Systems}} \bibinfo{volume}{23}, \bibinfo{number}{9} (\bibinfo{year}{2022}), \bibinfo{pages}{15808--15823}.
\newblock


\bibitem[Lv et~al\mbox{.}(2020)]%
        {lowlight2}
\bibfield{author}{\bibinfo{person}{Feifan Lv}, \bibinfo{person}{Bo Liu}, {and} \bibinfo{person}{Feng Lu}.} \bibinfo{year}{2020}\natexlab{}.
\newblock \showarticletitle{Fast enhancement for non-uniform illumination images using light-weight CNNs}. In \bibinfo{booktitle}{\emph{ACM MM}}. \bibinfo{pages}{1450--1458}.
\newblock


\bibitem[Ma et~al\mbox{.}(2015)]%
        {mefl}
\bibfield{author}{\bibinfo{person}{Kede Ma}, \bibinfo{person}{Kai Zeng}, {and} \bibinfo{person}{Zhou Wang}.} \bibinfo{year}{2015}\natexlab{}.
\newblock \showarticletitle{Perceptual quality assessment for multi-exposure image fusion}.
\newblock \bibinfo{journal}{\emph{IEEE TIP}} \bibinfo{volume}{24}, \bibinfo{number}{11} (\bibinfo{year}{2015}), \bibinfo{pages}{3345--3356}.
\newblock


\bibitem[Ma et~al\mbox{.}(2022)]%
        {lowlight5}
\bibfield{author}{\bibinfo{person}{Long Ma}, \bibinfo{person}{Tengyu Ma}, \bibinfo{person}{Risheng Liu}, \bibinfo{person}{Xin Fan}, {and} \bibinfo{person}{Zhongxuan Luo}.} \bibinfo{year}{2022}\natexlab{}.
\newblock \showarticletitle{Toward fast, flexible, and robust low-light image enhancement}. In \bibinfo{booktitle}{\emph{CVPR}}. \bibinfo{pages}{5637--5646}.
\newblock


\bibitem[Mittal et~al\mbox{.}(2013)]%
        {6353522}
\bibfield{author}{\bibinfo{person}{Anish Mittal}, \bibinfo{person}{Rajiv Soundararajan}, {and} \bibinfo{person}{Alan~C. Bovik}.} \bibinfo{year}{2013}\natexlab{}.
\newblock \showarticletitle{Making a “Completely Blind” Image Quality Analyzer}.
\newblock \bibinfo{journal}{\emph{IEEE Signal Processing Letters}} \bibinfo{volume}{20}, \bibinfo{number}{3} (\bibinfo{year}{2013}), \bibinfo{pages}{209--212}.
\newblock
\urldef\tempurl%
\url{https://doi.org/10.1109/LSP.2012.2227726}
\showDOI{\tempurl}


\bibitem[Nyberg et~al\mbox{.}(2018)]%
        {genfra1}
\bibfield{author}{\bibinfo{person}{Adam Nyberg}, \bibinfo{person}{Abdelrahman Eldesokey}, \bibinfo{person}{David Bergstrom}, {and} \bibinfo{person}{David Gustafsson}.} \bibinfo{year}{2018}\natexlab{}.
\newblock \showarticletitle{Unpaired thermal to visible spectrum transfer using adversarial training}. In \bibinfo{booktitle}{\emph{ECCV}}. \bibinfo{pages}{0--0}.
\newblock


\bibitem[Panetta et~al\mbox{.}(2016)]%
        {7305804}
\bibfield{author}{\bibinfo{person}{Karen Panetta}, \bibinfo{person}{Chen Gao}, {and} \bibinfo{person}{Sos Agaian}.} \bibinfo{year}{2016}\natexlab{}.
\newblock \showarticletitle{Human-Visual-System-Inspired Underwater Image Quality Measures}.
\newblock \bibinfo{journal}{\emph{IEEE Journal of Oceanic Engineering}} \bibinfo{volume}{41}, \bibinfo{number}{3} (\bibinfo{year}{2016}), \bibinfo{pages}{541--551}.
\newblock
\urldef\tempurl%
\url{https://doi.org/10.1109/JOE.2015.2469915}
\showDOI{\tempurl}


\bibitem[Park et~al\mbox{.}(2008)]%
        {histogram3}
\bibfield{author}{\bibinfo{person}{Gyu-Hee Park}, \bibinfo{person}{Hwa-Hyun Cho}, {and} \bibinfo{person}{Myung-Ryul Choi}.} \bibinfo{year}{2008}\natexlab{}.
\newblock \showarticletitle{A contrast enhancement method using dynamic range separate histogram equalization}.
\newblock \bibinfo{journal}{\emph{IEEE Transactions on Consumer Electronics}} \bibinfo{volume}{54}, \bibinfo{number}{4} (\bibinfo{year}{2008}), \bibinfo{pages}{1981--1987}.
\newblock


\bibitem[Richter et~al\mbox{.}(2017)]%
        {Richter_2017_ICCV}
\bibfield{author}{\bibinfo{person}{Stephan~R. Richter}, \bibinfo{person}{Zeeshan Hayder}, {and} \bibinfo{person}{Vladlen Koltun}.} \bibinfo{year}{2017}\natexlab{}.
\newblock \showarticletitle{Playing for Benchmarks}. In \bibinfo{booktitle}{\emph{ICCV}}.
\newblock


\bibitem[Sigillo et~al\mbox{.}(2023)]%
        {stawgan}
\bibfield{author}{\bibinfo{person}{Luigi Sigillo}, \bibinfo{person}{Eleonora Grassucci}, {and} \bibinfo{person}{Danilo Comminiello}.} \bibinfo{year}{2023}\natexlab{}.
\newblock \showarticletitle{StawGAN: Structural-aware generative adversarial networks for infrared image translation}. In \bibinfo{booktitle}{\emph{2023 IEEE International Symposium on Circuits and Systems (ISCAS)}}. IEEE, \bibinfo{pages}{1--5}.
\newblock


\bibitem[Simonyan and Zisserman(2014)]%
        {simonyan2014very}
\bibfield{author}{\bibinfo{person}{Karen Simonyan} {and} \bibinfo{person}{Andrew Zisserman}.} \bibinfo{year}{2014}\natexlab{}.
\newblock \showarticletitle{Very deep convolutional networks for large-scale image recognition}.
\newblock \bibinfo{journal}{\emph{arXiv preprint arXiv:1409.1556}} (\bibinfo{year}{2014}).
\newblock


\bibitem[Sun et~al\mbox{.}(2022)]%
        {sun2022shift}
\bibfield{author}{\bibinfo{person}{Tao Sun}, \bibinfo{person}{Mattia Segu}, \bibinfo{person}{Janis Postels}, \bibinfo{person}{Yuxuan Wang}, \bibinfo{person}{Luc Van~Gool}, \bibinfo{person}{Bernt Schiele}, \bibinfo{person}{Federico Tombari}, {and} \bibinfo{person}{Fisher Yu}.} \bibinfo{year}{2022}\natexlab{}.
\newblock \showarticletitle{SHIFT: a synthetic driving dataset for continuous multi-task domain adaptation}. In \bibinfo{booktitle}{\emph{CVPR}}. \bibinfo{pages}{21371--21382}.
\newblock


\bibitem[Suvorov et~al\mbox{.}(2022)]%
        {fourconvolution}
\bibfield{author}{\bibinfo{person}{Roman Suvorov}, \bibinfo{person}{Elizaveta Logacheva}, \bibinfo{person}{Anton Mashikhin}, \bibinfo{person}{Anastasia Remizova}, \bibinfo{person}{Arsenii Ashukha}, \bibinfo{person}{Aleksei Silvestrov}, \bibinfo{person}{Naejin Kong}, \bibinfo{person}{Harshith Goka}, \bibinfo{person}{Kiwoong Park}, {and} \bibinfo{person}{Victor Lempitsky}.} \bibinfo{year}{2022}\natexlab{}.
\newblock \showarticletitle{Resolution-robust large mask inpainting with fourier convolutions}. In \bibinfo{booktitle}{\emph{WACV}}. \bibinfo{pages}{2149--2159}.
\newblock


\bibitem[Wang et~al\mbox{.}(2023a)]%
        {four1}
\bibfield{author}{\bibinfo{person}{Chenxi Wang}, \bibinfo{person}{Hongujun Wu}, {and} \bibinfo{person}{Jin Zhi}.} \bibinfo{year}{2023}\natexlab{a}.
\newblock \showarticletitle{FourLLIE: Boosting Low-Light Image Enhancement by Fourier Frequency Information}. In \bibinfo{booktitle}{\emph{ACM MM}}.
\newblock


\bibitem[Wang et~al\mbox{.}(2013)]%
        {npe}
\bibfield{author}{\bibinfo{person}{Shuhang Wang}, \bibinfo{person}{Jin Zheng}, \bibinfo{person}{Hai-Miao Hu}, {and} \bibinfo{person}{Bo Li}.} \bibinfo{year}{2013}\natexlab{}.
\newblock \showarticletitle{Naturalness preserved enhancement algorithm for non-uniform illumination images}.
\newblock \bibinfo{journal}{\emph{IEEE TIP}} \bibinfo{volume}{22}, \bibinfo{number}{9} (\bibinfo{year}{2013}), \bibinfo{pages}{3538--3548}.
\newblock


\bibitem[Wang et~al\mbox{.}(2023b)]%
        {lowlight7}
\bibfield{author}{\bibinfo{person}{Yufei Wang}, \bibinfo{person}{Yi Yu}, \bibinfo{person}{Wenhan Yang}, \bibinfo{person}{Lanqing Guo}, \bibinfo{person}{Lap-Pui Chau}, \bibinfo{person}{Alex~C Kot}, {and} \bibinfo{person}{Bihan Wen}.} \bibinfo{year}{2023}\natexlab{b}.
\newblock \showarticletitle{Exposurediffusion: Learning to expose for low-light image enhancement}. In \bibinfo{booktitle}{\emph{ICCV}}. \bibinfo{pages}{12438--12448}.
\newblock


\bibitem[Wang et~al\mbox{.}(2004)]%
        {SSIM}
\bibfield{author}{\bibinfo{person}{Zhou Wang}, \bibinfo{person}{Alan~C Bovik}, \bibinfo{person}{Hamid~R Sheikh}, {and} \bibinfo{person}{Eero~P Simoncelli}.} \bibinfo{year}{2004}\natexlab{}.
\newblock \showarticletitle{Image quality assessment: from error visibility to structural similarity}.
\newblock \bibinfo{journal}{\emph{IEEE TIP}} \bibinfo{volume}{13}, \bibinfo{number}{4} (\bibinfo{year}{2004}), \bibinfo{pages}{600--612}.
\newblock


\bibitem[Wei et~al\mbox{.}(2018)]%
        {lolv1}
\bibfield{author}{\bibinfo{person}{Chen Wei}, \bibinfo{person}{Wenjing Wang}, \bibinfo{person}{Wenhan Yang}, {and} \bibinfo{person}{Jiaying Liu}.} \bibinfo{year}{2018}\natexlab{}.
\newblock \showarticletitle{Deep retinex decomposition for low-light enhancement}.
\newblock \bibinfo{journal}{\emph{arXiv preprint arXiv:1808.04560}} (\bibinfo{year}{2018}).
\newblock


\bibitem[Wu et~al\mbox{.}(2023)]%
        {wu2023learning}
\bibfield{author}{\bibinfo{person}{Yuhui Wu}, \bibinfo{person}{Chen Pan}, \bibinfo{person}{Guoqing Wang}, \bibinfo{person}{Yang Yang}, \bibinfo{person}{Jiwei Wei}, \bibinfo{person}{Chongyi Li}, {and} \bibinfo{person}{Heng~Tao Shen}.} \bibinfo{year}{2023}\natexlab{}.
\newblock \showarticletitle{Learning Semantic-Aware Knowledge Guidance for Low-Light Image Enhancement}. In \bibinfo{booktitle}{\emph{CVPR}}. \bibinfo{pages}{1662--1671}.
\newblock


\bibitem[Xu et~al\mbox{.}(2020)]%
        {retinex3}
\bibfield{author}{\bibinfo{person}{Jun Xu}, \bibinfo{person}{Yingkun Hou}, \bibinfo{person}{Dongwei Ren}, \bibinfo{person}{Li Liu}, \bibinfo{person}{Fan Zhu}, \bibinfo{person}{Mengyang Yu}, \bibinfo{person}{Haoqian Wang}, {and} \bibinfo{person}{Ling Shao}.} \bibinfo{year}{2020}\natexlab{}.
\newblock \showarticletitle{Star: A structure and texture aware retinex model}.
\newblock \bibinfo{journal}{\emph{IEEE TIP}}  \bibinfo{volume}{29} (\bibinfo{year}{2020}), \bibinfo{pages}{5022--5037}.
\newblock


\bibitem[Xu et~al\mbox{.}(2022)]%
        {lowlight8}
\bibfield{author}{\bibinfo{person}{Xiaogang Xu}, \bibinfo{person}{Ruixing Wang}, \bibinfo{person}{Chi-Wing Fu}, {and} \bibinfo{person}{Jiaya Jia}.} \bibinfo{year}{2022}\natexlab{}.
\newblock \showarticletitle{SNR-aware low-light image enhancement}. In \bibinfo{booktitle}{\emph{CVPR}}. \bibinfo{pages}{17714--17724}.
\newblock


\bibitem[Xu et~al\mbox{.}(2023)]%
        {xu2023low}
\bibfield{author}{\bibinfo{person}{Xiaogang Xu}, \bibinfo{person}{Ruixing Wang}, {and} \bibinfo{person}{Jiangbo Lu}.} \bibinfo{year}{2023}\natexlab{}.
\newblock \showarticletitle{Low-light image enhancement via structure modeling and guidance}. In \bibinfo{booktitle}{\emph{CVPR}}. \bibinfo{pages}{9893--9903}.
\newblock


\bibitem[Yang et~al\mbox{.}(2024)]%
        {depthanything}
\bibfield{author}{\bibinfo{person}{Lihe Yang}, \bibinfo{person}{Bingyi Kang}, \bibinfo{person}{Zilong Huang}, \bibinfo{person}{Xiaogang Xu}, \bibinfo{person}{Jiashi Feng}, {and} \bibinfo{person}{Hengshuang Zhao}.} \bibinfo{year}{2024}\natexlab{}.
\newblock \showarticletitle{Depth Anything: Unleashing the Power of Large-Scale Unlabeled Data}. In \bibinfo{booktitle}{\emph{CVPR}}.
\newblock


\bibitem[Yang et~al\mbox{.}(2021)]%
        {lol}
\bibfield{author}{\bibinfo{person}{Wenhan Yang}, \bibinfo{person}{Wenjing Wang}, \bibinfo{person}{Haofeng Huang}, \bibinfo{person}{Shiqi Wang}, {and} \bibinfo{person}{Jiaying Liu}.} \bibinfo{year}{2021}\natexlab{}.
\newblock \showarticletitle{Sparse gradient regularized deep retinex network for robust low-light image enhancement}.
\newblock \bibinfo{journal}{\emph{IEEE TIP}}  \bibinfo{volume}{30} (\bibinfo{year}{2021}), \bibinfo{pages}{2072--2086}.
\newblock


\bibitem[Yang et~al\mbox{.}(2020)]%
        {yang2020advancing}
\bibfield{author}{\bibinfo{person}{Wenhan Yang}, \bibinfo{person}{Ye Yuan}, \bibinfo{person}{Wenqi Ren}, \bibinfo{person}{Jiaying Liu}, \bibinfo{person}{Walter~J Scheirer}, \bibinfo{person}{Zhangyang Wang}, \bibinfo{person}{Taiheng Zhang}, \bibinfo{person}{Qiaoyong Zhong}, \bibinfo{person}{Di Xie}, \bibinfo{person}{Shiliang Pu}, {et~al\mbox{.}}} \bibinfo{year}{2020}\natexlab{}.
\newblock \showarticletitle{Advancing image understanding in poor visibility environments: A collective benchmark study}.
\newblock \bibinfo{journal}{\emph{IEEE TIP}}  \bibinfo{volume}{29} (\bibinfo{year}{2020}), \bibinfo{pages}{5737--5752}.
\newblock


\bibitem[Yu et~al\mbox{.}(2022)]%
        {fournoise2}
\bibfield{author}{\bibinfo{person}{Hu Yu}, \bibinfo{person}{Naishan Zheng}, \bibinfo{person}{Man Zhou}, \bibinfo{person}{Jie Huang}, \bibinfo{person}{Zeyu Xiao}, {and} \bibinfo{person}{Feng Zhao}.} \bibinfo{year}{2022}\natexlab{}.
\newblock \showarticletitle{Frequency and spatial dual guidance for image dehazing}. In \bibinfo{booktitle}{\emph{ECCV}}. Springer, \bibinfo{pages}{181--198}.
\newblock


\bibitem[Yun et~al\mbox{.}(2022)]%
        {9981857}
\bibfield{author}{\bibinfo{person}{Seungsang Yun}, \bibinfo{person}{Minwoo Jung}, \bibinfo{person}{Jeongyun Kim}, \bibinfo{person}{Sangwoo Jung}, \bibinfo{person}{Younghun Cho}, \bibinfo{person}{Myung-Hwan Jeon}, \bibinfo{person}{Giseop Kim}, {and} \bibinfo{person}{Ayoung Kim}.} \bibinfo{year}{2022}\natexlab{}.
\newblock \showarticletitle{STheReO: Stereo Thermal Dataset for Research in Odometry and Mapping}. In \bibinfo{booktitle}{\emph{2022 IEEE/RSJ International Conference on Intelligent Robots and Systems (IROS)}}. \bibinfo{pages}{3857--3864}.
\newblock
\urldef\tempurl%
\url{https://doi.org/10.1109/IROS47612.2022.9981857}
\showDOI{\tempurl}


\bibitem[Zamir et~al\mbox{.}(2022)]%
        {restormer}
\bibfield{author}{\bibinfo{person}{Syed~Waqas Zamir}, \bibinfo{person}{Aditya Arora}, \bibinfo{person}{Salman Khan}, \bibinfo{person}{Munawar Hayat}, \bibinfo{person}{Fahad~Shahbaz Khan}, {and} \bibinfo{person}{Ming-Hsuan Yang}.} \bibinfo{year}{2022}\natexlab{}.
\newblock \showarticletitle{Restormer: Efficient transformer for high-resolution image restoration}. In \bibinfo{booktitle}{\emph{CVPR}}. \bibinfo{pages}{5728--5739}.
\newblock


\bibitem[Zamir et~al\mbox{.}(2020)]%
        {lowlight9}
\bibfield{author}{\bibinfo{person}{Syed~Waqas Zamir}, \bibinfo{person}{Aditya Arora}, \bibinfo{person}{Salman Khan}, \bibinfo{person}{Munawar Hayat}, \bibinfo{person}{Fahad~Shahbaz Khan}, \bibinfo{person}{Ming-Hsuan Yang}, {and} \bibinfo{person}{Ling Shao}.} \bibinfo{year}{2020}\natexlab{}.
\newblock \showarticletitle{Learning enriched features for real image restoration and enhancement}. In \bibinfo{booktitle}{\emph{ECCV}}. Springer, \bibinfo{pages}{492--511}.
\newblock


\bibitem[Zhang et~al\mbox{.}(2021b)]%
        {lowlight3}
\bibfield{author}{\bibinfo{person}{Fan Zhang}, \bibinfo{person}{Yu Li}, \bibinfo{person}{Shaodi You}, {and} \bibinfo{person}{Ying Fu}.} \bibinfo{year}{2021}\natexlab{b}.
\newblock \showarticletitle{Learning temporal consistency for low light video enhancement from single images}. In \bibinfo{booktitle}{\emph{CVPR}}. \bibinfo{pages}{4967--4976}.
\newblock


\bibitem[Zhang et~al\mbox{.}(2018)]%
        {LP}
\bibfield{author}{\bibinfo{person}{Richard Zhang}, \bibinfo{person}{Phillip Isola}, \bibinfo{person}{Alexei~A Efros}, \bibinfo{person}{Eli Shechtman}, {and} \bibinfo{person}{Oliver Wang}.} \bibinfo{year}{2018}\natexlab{}.
\newblock \showarticletitle{The unreasonable effectiveness of deep features as a perceptual metric}. In \bibinfo{booktitle}{\emph{CVPR}}. \bibinfo{pages}{586--595}.
\newblock


\bibitem[Zhang et~al\mbox{.}(2021a)]%
        {kind++}
\bibfield{author}{\bibinfo{person}{Yonghua Zhang}, \bibinfo{person}{Xiaojie Guo}, \bibinfo{person}{Jiayi Ma}, \bibinfo{person}{Wei Liu}, {and} \bibinfo{person}{Jiawan Zhang}.} \bibinfo{year}{2021}\natexlab{a}.
\newblock \showarticletitle{Beyond brightening low-light images}.
\newblock \bibinfo{journal}{\emph{IJCV}}  \bibinfo{volume}{129} (\bibinfo{year}{2021}), \bibinfo{pages}{1013--1037}.
\newblock


\bibitem[Zhang and Wang(2023)]%
        {zhang2023diverse}
\bibfield{author}{\bibinfo{person}{Yukang Zhang} {and} \bibinfo{person}{Hanzi Wang}.} \bibinfo{year}{2023}\natexlab{}.
\newblock \showarticletitle{Diverse Embedding Expansion Network and Low-Light Cross-Modality Benchmark for Visible-Infrared Person Re-identification}. In \bibinfo{booktitle}{\emph{CVPR}}. \bibinfo{pages}{2153--2162}.
\newblock


\bibitem[Zhang et~al\mbox{.}(2019)]%
        {kind}
\bibfield{author}{\bibinfo{person}{Yonghua Zhang}, \bibinfo{person}{Jiawan Zhang}, {and} \bibinfo{person}{Xiaojie Guo}.} \bibinfo{year}{2019}\natexlab{}.
\newblock \showarticletitle{Kindling the darkness: A practical low-light image enhancer}. In \bibinfo{booktitle}{\emph{ACM MM}}. \bibinfo{pages}{1632--1640}.
\newblock


\bibitem[Zhao et~al\mbox{.}(2023b)]%
        {fuse2}
\bibfield{author}{\bibinfo{person}{Wenda Zhao}, \bibinfo{person}{Shigeng Xie}, \bibinfo{person}{Fan Zhao}, \bibinfo{person}{You He}, {and} \bibinfo{person}{Huchuan Lu}.} \bibinfo{year}{2023}\natexlab{b}.
\newblock \showarticletitle{MetaFusion: Infrared and Visible Image Fusion via Meta-Feature Embedding From Object Detection}. In \bibinfo{booktitle}{\emph{CVPR}}. \bibinfo{pages}{13955--13965}.
\newblock


\bibitem[Zhao et~al\mbox{.}(2023a)]%
        {zhao2023cddfuse}
\bibfield{author}{\bibinfo{person}{Zixiang Zhao}, \bibinfo{person}{Haowen Bai}, \bibinfo{person}{Jiangshe Zhang}, \bibinfo{person}{Yulun Zhang}, \bibinfo{person}{Shuang Xu}, \bibinfo{person}{Zudi Lin}, \bibinfo{person}{Radu Timofte}, {and} \bibinfo{person}{Luc Van~Gool}.} \bibinfo{year}{2023}\natexlab{a}.
\newblock \showarticletitle{Cddfuse: Correlation-driven dual-branch feature decomposition for multi-modality image fusion}. In \bibinfo{booktitle}{\emph{CVPR}}. \bibinfo{pages}{5906--5916}.
\newblock


\bibitem[Zheng et~al\mbox{.}(2021)]%
        {lowlight4}
\bibfield{author}{\bibinfo{person}{Chuanjun Zheng}, \bibinfo{person}{Daming Shi}, {and} \bibinfo{person}{Wentian Shi}.} \bibinfo{year}{2021}\natexlab{}.
\newblock \showarticletitle{Adaptive unfolding total variation network for low-light image enhancement}. In \bibinfo{booktitle}{\emph{ICCV}}. \bibinfo{pages}{4439--4448}.
\newblock


\bibitem[Zheng et~al\mbox{.}(2023)]%
        {lowlight6}
\bibfield{author}{\bibinfo{person}{Naishan Zheng}, \bibinfo{person}{Man Zhou}, \bibinfo{person}{Yanmeng Dong}, \bibinfo{person}{Xiangyu Rui}, \bibinfo{person}{Jie Huang}, \bibinfo{person}{Chongyi Li}, {and} \bibinfo{person}{Feng Zhao}.} \bibinfo{year}{2023}\natexlab{}.
\newblock \showarticletitle{Empowering Low-Light Image Enhancer through Customized Learnable Priors}. In \bibinfo{booktitle}{\emph{ICCV}}. \bibinfo{pages}{12559--12569}.
\newblock


\bibitem[Zhou et~al\mbox{.}(2022)]%
        {fournoise1}
\bibfield{author}{\bibinfo{person}{Man Zhou}, \bibinfo{person}{Jie Huang}, \bibinfo{person}{Keyu Yan}, \bibinfo{person}{Hu Yu}, \bibinfo{person}{Xueyang Fu}, \bibinfo{person}{Aiping Liu}, \bibinfo{person}{Xian Wei}, {and} \bibinfo{person}{Feng Zhao}.} \bibinfo{year}{2022}\natexlab{}.
\newblock \showarticletitle{Spatial-frequency domain information integration for pan-sharpening}. In \bibinfo{booktitle}{\emph{ECCV}}. Springer, \bibinfo{pages}{274--291}.
\newblock


\end{thebibliography}

\appendix

\end{document}